\PassOptionsToPackage{table}{xcolor}
\documentclass{article} 
\usepackage[T1]{fontenc}
\usepackage{iclr2027_conference,times}
\iclrfinalcopy

\usepackage{amsmath,amsfonts,bm}

\def\eqref#1{equation~\ref{#1}}

\def\1{\bm{1}}

\DeclareMathAlphabet{\mathsfit}{\encodingdefault}{\sfdefault}{m}{sl}
\SetMathAlphabet{\mathsfit}{bold}{\encodingdefault}{\sfdefault}{bx}{n}

\usepackage{graphicx}
\usepackage{float}
\usepackage{wrapfig}
\usepackage{multirow}
\usepackage{hyperref}
\usepackage{url}
\usepackage{booktabs}
\usepackage[table]{xcolor}
\usepackage{pifont}
\usepackage{fontawesome5}
\usepackage[most]{tcolorbox}
\tcbuselibrary{listings,breakable}
\newtcblisting{promptbox}[1]{enhanced,breakable,listing only,listing engine=listings,
  colback=blue!2,colframe=teal!65!black,colbacktitle=teal!78!black,
  coltitle=white,title={#1},fonttitle=\bfseries\small,
  boxrule=0.55pt,arc=2pt,left=5pt,right=5pt,top=4pt,bottom=4pt,
  listing options={basicstyle=\ttfamily\footnotesize,columns=fullflexible,
    breaklines=true,breakatwhitespace=true,keepspaces=true,showstringspaces=false}}
\newtcblisting{recordbox}[1]{enhanced,breakable,listing only,listing engine=listings,
  colback=blue!3,colframe=blue!38!black,colbacktitle=blue!58!black,
  coltitle=white,title={#1},fonttitle=\bfseries\small,
  boxrule=0.55pt,arc=2pt,left=5pt,right=5pt,top=2pt,bottom=2pt,
  listing options={basicstyle=\ttfamily\fontsize{7.5}{8.5}\selectfont,columns=fullflexible,
    breaklines=true,breakatwhitespace=false,keepspaces=true,showstringspaces=false}}

\definecolor{APMInsightBlue}{HTML}{DCECF8}
\definecolor{APMInsightGreen}{HTML}{E0EFE6}
\definecolor{APMInsightPurple}{HTML}{EBE3F5}
\definecolor{APMInsightYellow}{HTML}{FDF3D9}
\newtcolorbox{insightbox}[2]{enhanced,
  colback=#1!35!white,colframe=#1!85!black,colbacktitle=#1,
  coltitle=black,title={#2},fonttitle=\normalfont\bfseries,
  fontupper=\normalfont\normalsize\itshape,
  boxrule=0.5pt,arc=2pt,left=6pt,right=6pt,top=3pt,bottom=3pt,
  toptitle=2pt,bottomtitle=2pt,before skip=6pt,
  after={\par\nobreak\smallskip}}

\newlength{\arxivOriginalIntextSep}
\title{\centering\raisebox{-0.03em}{\includegraphics[width=1.8em,trim=20bp 364bp 20bp 400bp,clip]{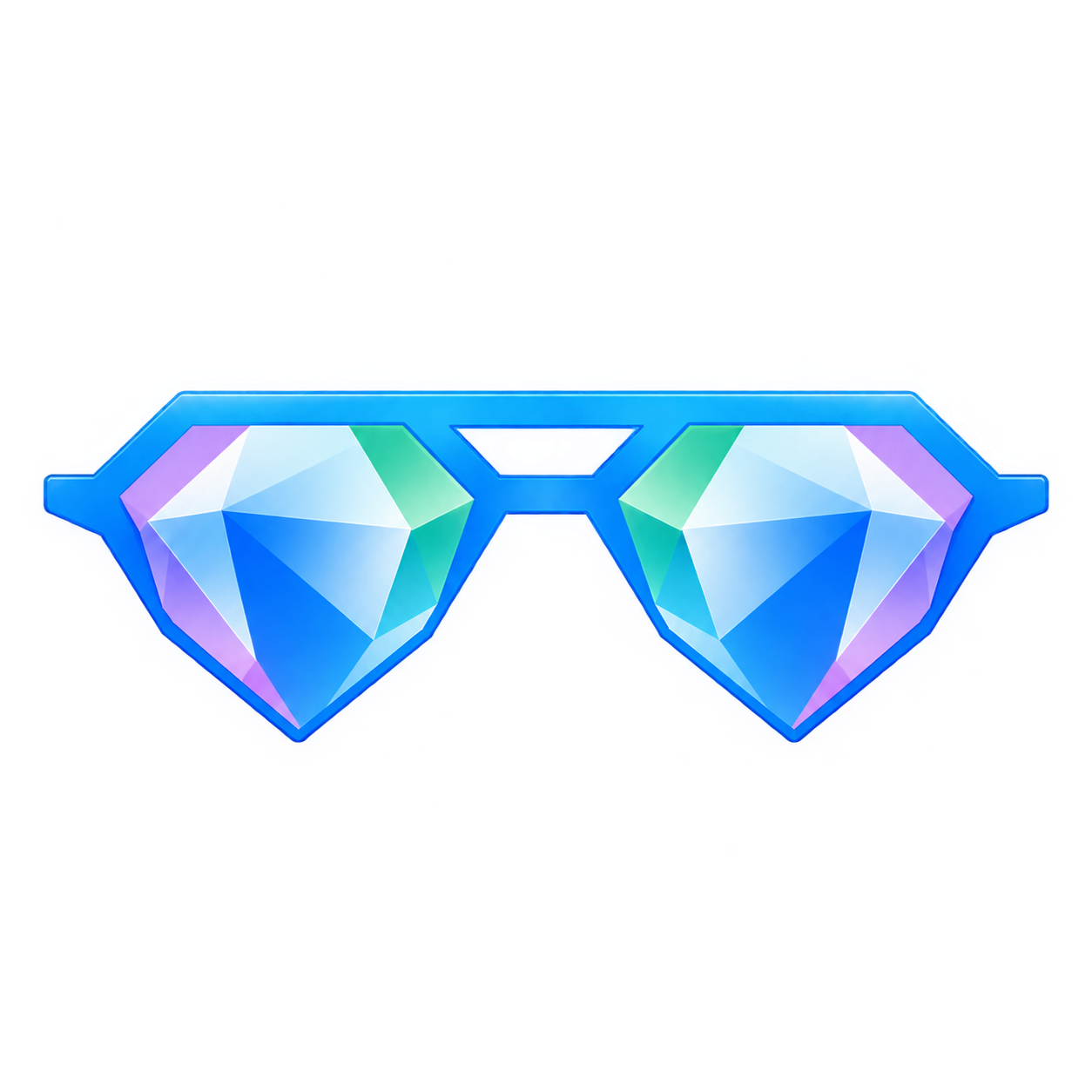}}\hspace{0.25em}\textcolor[HTML]{44ACFF}{A}\textcolor[HTML]{26B978}{P}\textcolor[HTML]{B987FF}{M}-Bench: Benchm\textcolor[HTML]{44ACFF}{a}rking cross-Session \textcolor[HTML]{26B978}{P}ersistent \textcolor[HTML]{B987FF}{M}emory for Real-World Egocentric Streaming Video Assistants}

\author{\begin{minipage}[t]{\dimexpr\textwidth-2\tabcolsep\relax}
\centering\normalfont\fontsize{11}{13}\selectfont
\makebox[\linewidth][c]{%
\textbf{Jianguo Huang}\textsuperscript{1,2}\thanks{Equal contribution.}\hspace{0.7em}%
\textbf{Jinming Liu}\textsuperscript{1,2}\footnotemark[1]\hspace{0.7em}%
\textbf{Qiyao Wang}\textsuperscript{3}\hspace{0.7em}%
\textbf{Liang Xu}\textsuperscript{4}\hspace{0.7em}%
\textbf{Jianhang Li}\textsuperscript{5}\hspace{0.7em}%
\textbf{Zhimian Wen}\textsuperscript{2}}\\
\makebox[\linewidth][c]{%
\textbf{Mingda Li}\textsuperscript{5}\hspace{0.7em}%
\textbf{Shule Lu}\textsuperscript{6}\hspace{0.7em}%
\textbf{Zhicheng Wang}\textsuperscript{2,7}\hspace{0.7em}%
\textbf{Yuhan Guo}\textsuperscript{1,2}\hspace{0.7em}%
\textbf{Xin Jin}\textsuperscript{2}\hspace{0.7em}%
\textbf{Wenjun Zeng}\textsuperscript{2}\thanks{Corresponding author.}\hspace{0.35em}}\\[0.7em]
\fontsize{11}{13}\selectfont
\mbox{\textsuperscript{1}Shanghai Jiao Tong University}\quad
\mbox{\textsuperscript{2}Eastern Institute of Technology, Ningbo}\\
\mbox{\textsuperscript{3}Shenzhen Institutes of Advanced Technology, Chinese Academy of Sciences}\\
\mbox{\textsuperscript{4}Zhongguancun Academy, Beijing, China}\quad
\mbox{\textsuperscript{5}Dalian University of Technology}\\
\makebox[\linewidth][c]{%
\textsuperscript{6}Beihang University\hspace{0.5em}%
\textsuperscript{7}Hong Kong Polytechnic University}\\[0.7em]
\makebox[\linewidth][c]{%
\normalfont\fontsize{10}{12}\selectfont
\href{https://github.com/Jianguo-Huang11/APM-Bench/tree/main}{\faGithub\enspace Code}\qquad
\href{https://jianguo-huang11.github.io/APM-Bench/}{\faGlobe\enspace Website}\qquad
\href{https://huggingface.co/datasets/Jianguo-Huang11/APM-Bench}{\faDatabase\enspace Dataset}}
\end{minipage}
}

\begin{document}

\null\kern-48pt
\maketitle
\fancyhead{}
\renewcommand{\headrulewidth}{0.4pt}
\thispagestyle{plain}
\raggedbottom

\begin{abstract}
To serve as real-world personal assistants, streaming video models need persistent memory that retains past experiences for later use. Yet existing streaming benchmarks and methods often focus on individual continuous videos or short clips, overlooking that real-world interactions are often intermittent and require memory to persist across interruptions.
To fill this gap, we introduce APM-Bench, which reformulates real-world streaming interaction as multi-session life trajectories. It contains 549 sessions, 104 trajectories, and 2,719 candidates, spanning both objective and open-ended questions. Each session is a video with fine-grained annotations, and sessions within a trajectory revolve around related activities. Models then use persistent memory to answer questions about past sessions and provide proactive responses while maintaining real-time interaction. This raises challenges: persistent memory must be storable, selectively retain information, be injected at the right time, and remain efficient. Moreover, finite storage may leave required evidence unavailable, so assistants should recognize missing evidence. Therefore, we systematically evaluate general video models under different memory protocols and diverse specialized streaming memory systems, and test whether models acknowledge insufficient evidence. Our evaluation reveals a clear utility--latency--storage trade-off: existing methods still struggle to simultaneously achieve reliable long-term recall, low overhead, and effective proactive assistance across sessions. APM-Bench provides a comprehensive testbed for developing and comparing persistent memory systems under realistic streaming conditions. We hope it encourages future work that jointly considers utility, latency, and storage toward more practical persistent memory for real-world streaming assistants.

\end{abstract}

\section{Introduction}

Streaming video models increasingly support continuous perception, real-time interaction, and proactive assistance~\citep{arxiv260614777,arxiv260913814}, showing their potential as personal assistants; memory is key to making such assistants truly personal by retaining and reusing user-specific experience over time. However, most existing benchmarks and methods study memory within a single continuous video, typically over a limited time span. In the real world, interactions are intermittent: users may turn off smart glasses and resume using the assistant hours or days later. The assistant must therefore retain and use relevant visual evidence from earlier interactions to answer later questions and provide proactive assistance.

\begin{figure}[t]
  \centering
  \includegraphics[width=\linewidth]{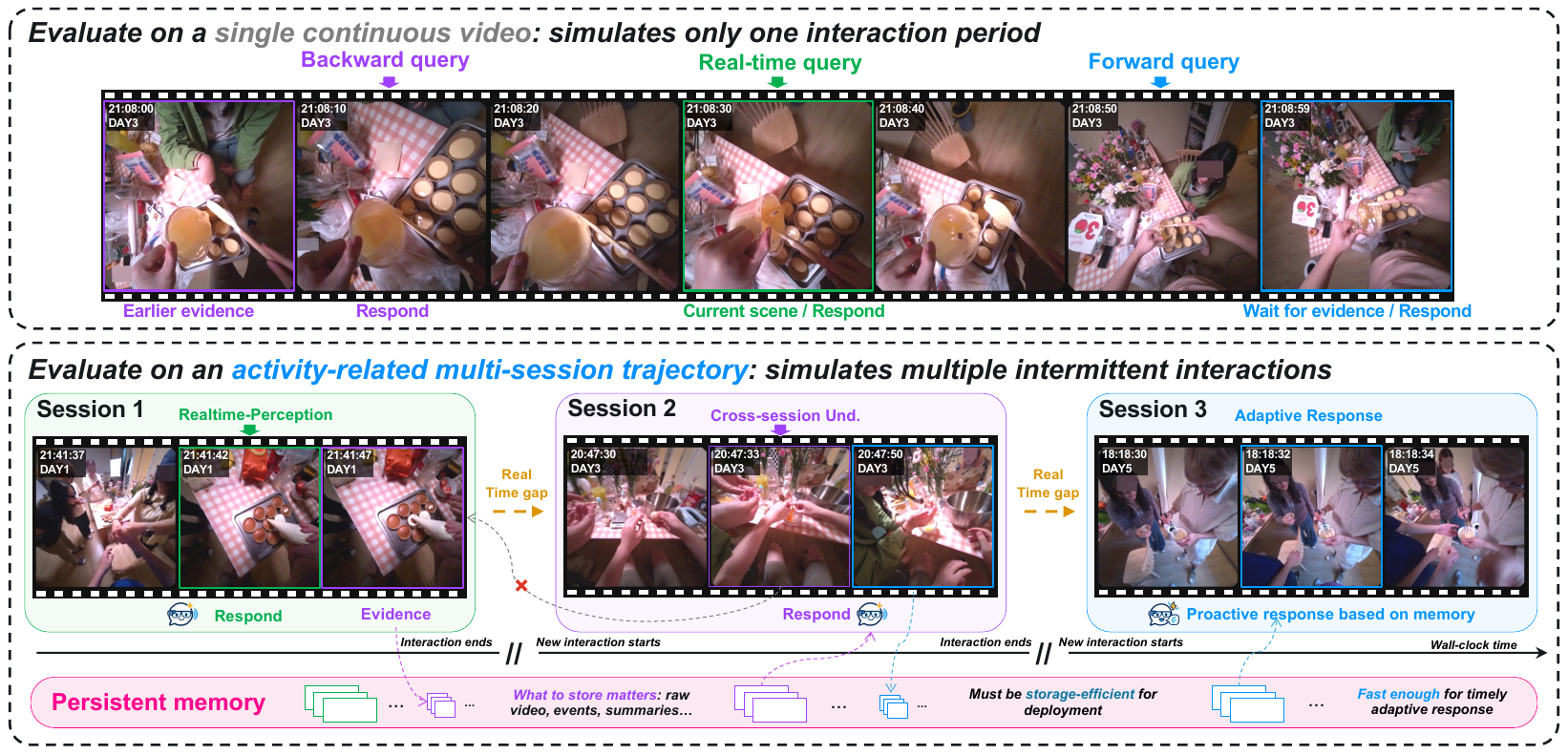}
  \caption{In a streaming setting, once an interaction ends, the model can no longer directly access its visual stream because real-world interactions are not replayed; the assistant therefore needs persistent memory to retain prior experience. Across sessions, the assistant updates and reuses persistent memory for cross-session understanding and proactive assistance while continuing real-time perception. The example shows repeated collaborative dessert-making across multiple sessions.}
  \label{fig:teaser}
\end{figure}

This calls for \textbf{persistent memory}: a storable record of past experience that remains available after an interaction ends and can be reused in later interactions. For real-world assistants, \textbf{\emph{such memory must support later tasks while keeping storage and response latency manageable}}. Yet existing evaluations rarely assess memory utility together with these deployment costs. Streaming video benchmarks evaluate understanding within individual videos~\citep{arxiv250105510,arxiv241103628} and extend interaction to longer continuous streams~\citep{arxiv260805703}. At much longer timescales, benchmarks assess streaming episodic memory~\citep{arxiv260531557} and long-term proactive service~\citep{arxiv260711523}, while proactive interaction benchmarks evaluate when models should respond and what assistance they should provide~\citep{arxiv251014560,arxiv260507299,arxiv260518577}. Overall, previous evaluations do not yet provide a clear picture of how persistent memory supports retrospective understanding and proactive assistance across temporally separated interactions while balancing utility, storage, and response latency.


\setlength{\intextsep}{-4pt}
\begin{wrapfigure}{r}{0.49\textwidth}
  \centering
  \includegraphics[width=\linewidth]{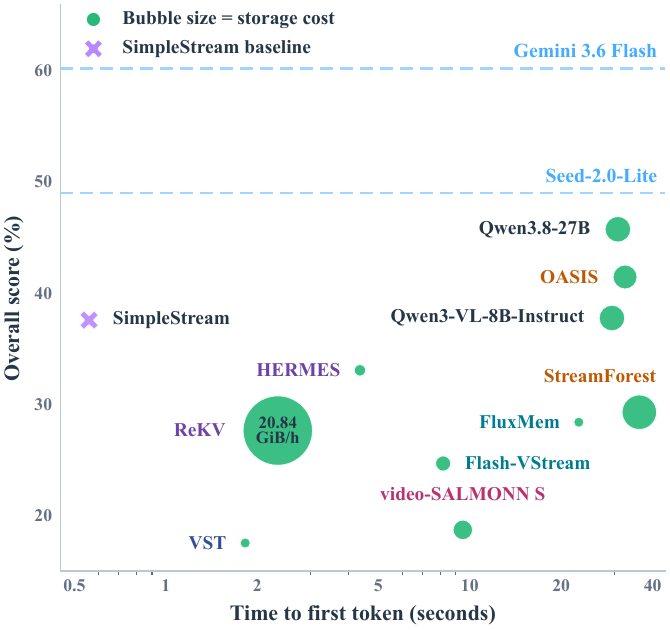}
    \caption{Utility-latency-storage trade-off across evaluated methods. Bubble size represents storage cost per hour. All general video models are evaluated under \emph{Raw Video as Memory}.}
  \label{fig:tradeoff}
\end{wrapfigure}

\begin{table}[t]
  \caption{Comparison of streaming video benchmarks. Most prior benchmarks evaluate models on a single continuous video; APM-Bench evaluates interactions across sessions separated by interruptions, simulating intermittent real-world use and testing whether memory from earlier sessions remains useful. It also tests whether models recognize when required historical evidence is unavailable. RTP: real-time perception; RET: retrospective tasks; PRO: proactive response; OE: open-ended candidates; OBJ: objective candidates; SE: storage-efficiency evaluation; MS: multi-session evaluation across related activities; EA: Evidence Availability-Aware evaluation.}
  \label{tab:benchmark-comparison}
  \centering
  \newcommand{\benchcheck}{\textcolor[HTML]{16A596}{\ding{51}}}
  \newcommand{\benchcross}{\textcolor{gray!65}{\ding{55}}}
  \setlength{\tabcolsep}{5pt}
  \renewcommand{\arraystretch}{1.12}
  \small
  \begin{tabular}{lcccccccc}
    \toprule
    \textbf{Benchmark} & \textbf{RTP} & \textbf{RET} & \textbf{PRO} & \textbf{OE} & \textbf{OBJ} & \textbf{SE} & \textbf{MS} & \textbf{EA} \\
    \midrule
    StreamingBench~\citep{arxiv241103628} & \benchcheck & \benchcheck & \benchcheck & \benchcross & \benchcheck & \benchcross & \benchcross & \benchcross \\
    OVO-Bench~\citep{arxiv250105510} & \benchcheck & \benchcheck & \benchcheck & \benchcross & \benchcheck & \benchcross & \benchcross & \benchcross \\
    PhoStream~\citep{arxiv260122575} & \benchcheck & \benchcheck & \benchcheck & \benchcheck & \benchcross & \benchcross & \benchcross & \benchcross \\
    EgoStream~\citep{arxiv260531557} & \benchcheck & \benchcheck & \benchcross & \benchcross & \benchcheck & \benchcheck & \benchcross & \benchcross \\
    EgoServe~\citep{arxiv260711523} & \benchcross & \benchcross & \benchcheck & \benchcheck & \benchcross & \benchcross & \benchcross & \benchcross \\
    StreamArena~\citep{arxiv260805703} & \benchcheck & \benchcheck & \benchcheck & \benchcheck & \benchcross & \benchcross & \benchcross & \benchcross \\
    \midrule
    \rowcolor{blue!5}
    \textbf{APM-Bench (Ours)} & \benchcheck & \benchcheck & \benchcheck & \benchcheck & \benchcheck & \benchcheck & \benchcheck & \benchcheck \\
    \bottomrule
  \end{tabular}
\end{table}

\begin{figure}[t]
  \centering
  \includegraphics[width=\linewidth]{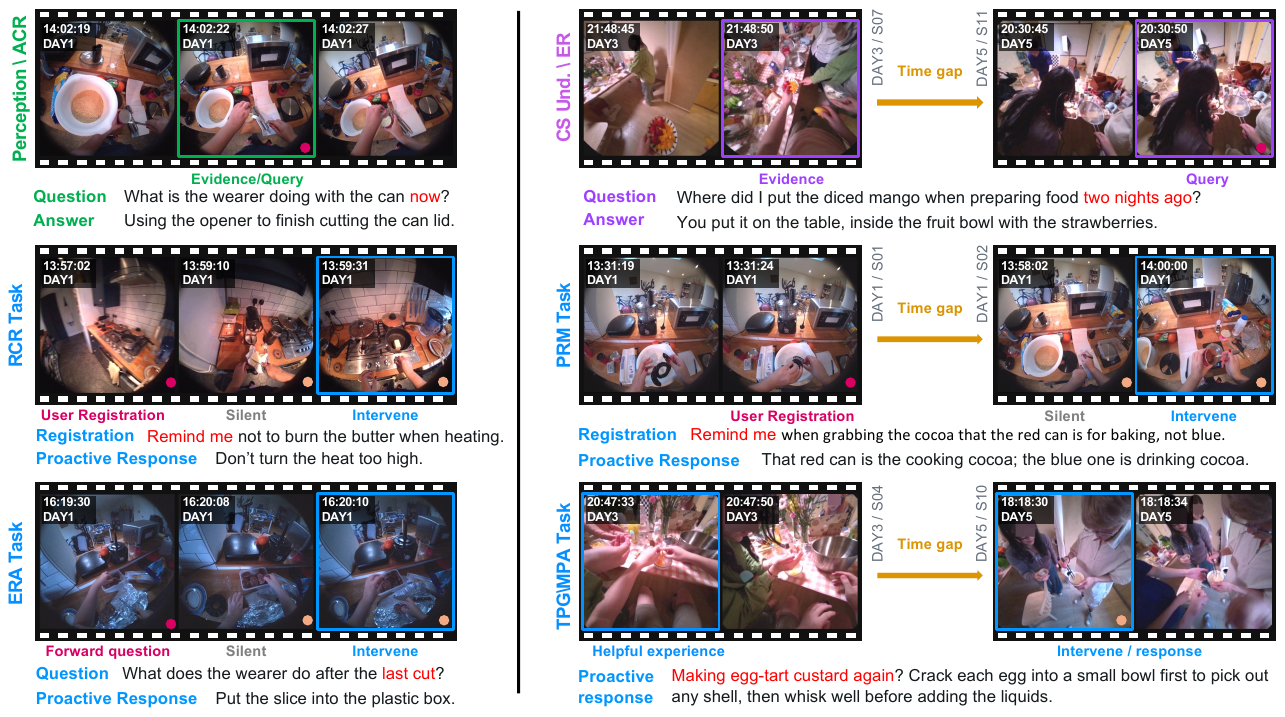}
  \caption{\textbf{Examples of intra-session and inter-session streaming tasks.} Each session follows its original continuous wall-clock timeline. Across the 12 tasks, we characterize six streaming temporal formulations: \emph{Cross-session understanding} and \emph{Real-time Perception} each follow a unified setup, while \emph{Adaptive Response} tasks adopt four distinct patterns (with TPG and MPA sharing one). \textbf{Notations:} {\color[HTML]{C72A64}$\bullet$} denotes user instructions, such as forward questions or reminder registrations; TPG and MPA trigger autonomously without explicit user prompts. {\color[HTML]{E7AD8A}$\bullet$} denotes probes used to evaluate whether the model should remain \texttt{SILENT} or \texttt{INTERVENE}. \emph{{\color[HTML]{4EAD5B}Real-time Perception}}: evidence and query co-occur in the current session; \emph{{\color[HTML]{93358F}Cross-session Understanding}}: evidence comes from prior sessions; and \emph{{\color[HTML]{3F8DF7}Adaptive Response}}: evidence comes from the current session for intra-session tasks (ERA and RCR), or from prior sessions for inter-session tasks (MPA, PRM, and TPG).
}
  \label{fig:task-examples}
\end{figure}

\begin{figure}[t]
  \centering
  \includegraphics[width=\linewidth]{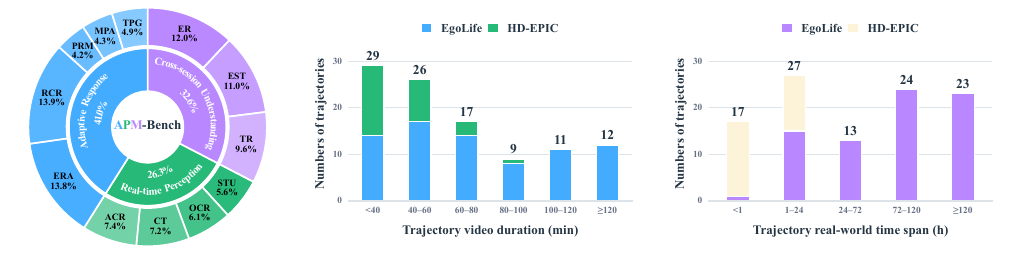}
  \caption{\textbf{APM-Bench statistics.} Left: distribution of 2{,}719 candidates across 12 tasks and three capability families. Middle: cumulative video duration across sessions in each trajectory. Right: wall-clock span from the first session start to the last session end, including inter-session gaps.}
  \label{fig:benchmark-distributions}
\end{figure}

\begin{figure}[t]
  \centering
  \includegraphics[width=\linewidth]{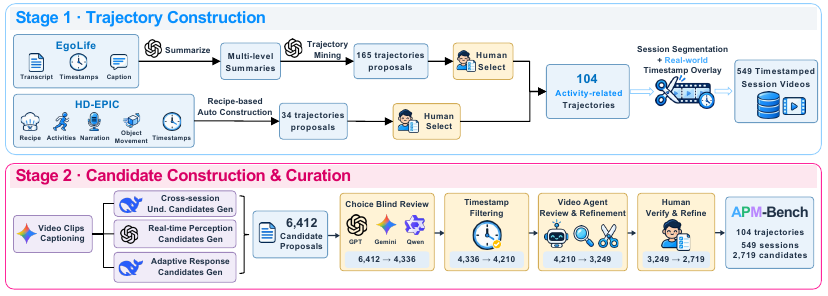}
  \caption{\textbf{APM-Bench construction pipeline.} In \emph{Stage 1}, EgoLife and HD-EPIC are organized into activity-related trajectories and segmented into sessions; real-world timestamps from the source metadata are rendered onto session videos that lack visible timestamps. In \emph{Stage 2}, candidates are generated from the session videos and progressively filtered and refined through choice-blind review, timestamp filtering, video-agent review, and human verification, yielding 2{,}719 candidates across 104 trajectories and 549 sessions. See Appendix~\ref{app:dataset-construction} for details.}
  \label{fig:benchmark-construction}
\end{figure}

\begin{table}[t]
\caption{Evaluation metrics across capability families and the \emph{Evidence Availability-Aware} setting.}
\label{tab:evaluation-metrics}
\centering
\small
\setlength{\tabcolsep}{7pt}
\renewcommand{\arraystretch}{1.08}
\begin{tabular}{lcc}
\toprule
\textbf{Evaluation} & \textbf{Output Format} & \textbf{Metric} \\
\midrule
\emph{Cross-session Understanding} & 4-way MCQA & Accuracy \\
\emph{Real-time Perception} & 4-way MCQA & Accuracy \\
\emph{Evidence Availability-Aware} & 5-way MCQA & Accuracy \\
\emph{Adaptive Response} & Open-ended & Gated LLM-Judge \\
\bottomrule
\end{tabular}
\end{table}

\begin{table}[t]
  \caption{Task performance and efficiency of the evaluated memory methods. Streaming and Persistent indicate support for streaming input and persistent memory, respectively. CS Und.: \emph{Cross-session Understanding}; Perception: \emph{Real-time Perception}; Adaptive: \emph{Adaptive Response}. Storage cost is reported per video hour, and Overall is the mean of the three capability scores.}
  \label{tab:efficiency-analysis}
  \centering
  \renewcommand{\arraystretch}{1.08}
  \setlength{\tabcolsep}{3.5pt}
  \newcommand{\effcheck}{\textcolor[HTML]{16A596}{\ding{51}}}
  \newcommand{\effcross}{\textcolor{gray!65}{\ding{55}}}
  \resizebox{\textwidth}{!}{%
  \begin{tabular}{@{}lcccccccc@{}}
    \toprule
    \textbf{Method}
      & \textbf{Streaming}
      & \textbf{Persistent}
      & \textbf{CS Und.}
      & \textbf{Perception}
      & \textbf{Adaptive}
      & \textbf{TTFT (s)}
      & \textbf{Storage Cost}
      & \textbf{Overall} \\
    \midrule
    \rowcolor{gray!15}
    \multicolumn{9}{l}{\textit{\textbf{w/o Memory}}} \\
    SimpleStream (Recent-4) & \effcross & \effcross & 27.21 & 53.06 & 32.48 & 0.56 & -- & 37.58 \\
    \specialrule{0.6pt}{0.4ex}{0pt}
    \specialrule{0.6pt}{0.4ex}{0.6ex}
    \rowcolor{gray!15}
    \multicolumn{9}{l}{\textit{\textbf{Raw Video as Memory}}} \\
    Seed-2.0-Lite~\citep{seed2026modelcard} & -- & -- & \underline{54.47} & 50.75 & 41.88 & -- & \multirow{3}{*}{3.01 GiB} & 49.03 \\
    Gemini 3.6 Flash~\citep{gemini36flash2026} & -- & -- & \textbf{69.37} & \underline{64.20} & \underline{47.05} & -- &  & \textbf{60.21} \\
    Qwen3.8-27B~\citep{qwen3827b2026} & -- & -- & 50.10 & 48.84 & 38.31 & 30.72 &  & 45.75 \\
    \midrule
    \rowcolor{gray!15}
    \multicolumn{9}{l}{\textit{\textbf{Text Summary as Memory}}} \\
    Seed-2.0-Lite~\citep{seed2026modelcard} & -- & -- & 45.34 & 52.44 & 44.08 & -- & 7.75 KiB & 47.29 \\
    Gemini 3.6 Flash~\citep{gemini36flash2026} & -- & -- & 47.50 & \textbf{65.98} & \textbf{47.14} & -- & 3.09 KiB & \underline{53.54} \\
    Qwen3.8-27B~\citep{qwen3827b2026} & -- & -- & 41.05 & 48.19 & 37.90 & 8.13 & 6.84 KiB & 42.38 \\
    \specialrule{0.6pt}{0.4ex}{0pt}
    \specialrule{0.6pt}{0.4ex}{0.6ex}
    \rowcolor{gray!15}
    \multicolumn{9}{l}{\textit{\textbf{KV Cache}}} \\
    HERMES~\citep{hermes2026} & \effcheck & \effcross & 34.12 & 42.04 & \underline{23.05} & 4.35 & 0.67 GiB & \underline{33.07} \\
    ReKV~\citep{rekv2025} & \effcheck & \effcross & 31.24 & 32.90 & 18.80 & \underline{2.34} & 20.84 GiB & 27.65 \\
    \midrule
    \rowcolor{gray!15}
    \multicolumn{9}{l}{\textit{\textbf{Visual Tokens / Features}}} \\
    FLUXMem~\citep{fluxmem2026} & \effcross & \effcross & \underline{35.98} & 27.59 & 21.63 & 22.85 & \underline{0.33 GiB} & 28.40 \\
    Flash-VStream~\citep{flashvstream2025} & \effcross & \effcross & 31.89 & 30.76 & 11.41 & 8.17 & 1.13 GiB & 24.69 \\
    \midrule
    \rowcolor{gray!15}
    \multicolumn{9}{l}{\textit{\textbf{Event Tree}}} \\
    StreamForest~\citep{streamforest2025} & \effcross & \effcross & \textbf{37.79} & \underline{43.62} & 6.48 & 36.15 & 5.34 GiB & 29.30 \\
    OASIS~\citep{oasis2026} & \effcheck & \effcheck & 35.46 & \textbf{52.92} & \textbf{35.99} & 32.44 & 2.66 GiB & \textbf{41.46} \\
    \midrule
    \rowcolor{gray!15}
    \multicolumn{9}{l}{\textit{\textbf{Parametric Memory}}} \\
    Video-Salmon-S~\citep{arxiv251011129} & \effcross & \effcross & 15.17 & 22.87 & 18.12 & 9.49 & 1.81 GiB & 18.72 \\
    \midrule
    \rowcolor{gray!15}
    \multicolumn{9}{l}{\textit{\textbf{Reasoning Thoughts}}} \\
    VST~\citep{arxiv260312262} & \effcheck & \effcross & 20.63 & 16.75 & 15.24 & \textbf{1.82} & \textbf{11.64 KiB} & 17.54 \\
    \bottomrule
  \end{tabular}%
  }
\end{table}

\begin{table}[t]
  \caption{Task-level results across the three capability families and \emph{Evidence Availability-Aware evaluation}. EA: evidence available in accessible session. EU: evidence unavailable detection.}
  \label{tab:main-results}
  \centering
  \renewcommand{\arraystretch}{1.08}
  \setlength{\tabcolsep}{1.5pt}
  \newlength{\mlacolwidth}
  \settowidth{\mlacolwidth}{\textbf{Evidence Availability-Aware}}
  \setlength{\mlacolwidth}{0.5\mlacolwidth}
  \resizebox{\textwidth}{!}{%
  \begin{tabular}{@{}l@{\hspace{9pt}}ccc@{\hspace{10pt}}cccc@{\hspace{10pt}}ccccc@{\hspace{6pt}}c@{\hspace{6pt}}*{2}{>{\centering\arraybackslash}p{\mlacolwidth}}@{}}
    \toprule
    \multirow{2}{*}{\textbf{Model / Method}}
      & \multicolumn{3}{c}{\hspace{-2pt}\textbf{Cross-session Und.}\hspace{2pt}}
      & \multicolumn{4}{c}{\hspace{-3pt}\textbf{Real-time Perception}\hspace{3pt}}
      & \multicolumn{5}{c}{\hspace{-3pt}\textbf{Adaptive Response}\hspace{3pt}}
      & \multirow{2}{*}{\textbf{Overall}}
      & \multicolumn{2}{c}{\textbf{Evidence Availability-Aware}} \\
    \cmidrule(l{3pt}r{7pt}){2-4}\cmidrule(l{2pt}r{8pt}){5-8}\cmidrule(l{2pt}r{8pt}){9-13}\cmidrule(lr){15-16}
      & \textbf{ER} & \textbf{EST} & \textbf{TR}
      & \textbf{ACR} & \textbf{CT} & \textbf{OCR} & \textbf{STU}
      & \textbf{ERA} & \textbf{RCR} & \textbf{MPA} & \textbf{PRM} & \textbf{TPG}
      & & \textbf{EA} & \textbf{EU} \\
    \midrule
    \rowcolor{gray!15}
    \multicolumn{16}{l}{\textit{\textbf{w/o Memory}}} \\
    SimpleStream & 27.52 & 28.52 & 25.57 & 56.44 & 35.90 & 73.49 & 46.41 & 50.89 & 30.22 & 26.54 & 32.11 & 22.64 & 37.58 & 3.85 & 95.38 \\
    \specialrule{0.6pt}{0.4ex}{0pt}
    \specialrule{0.6pt}{0.4ex}{0.6ex}
    \rowcolor{gray!15}
    \multicolumn{16}{l}{\textit{\textbf{Raw Video as Memory}}} \\
    Seed-2.0-Lite & 54.13 & \underline{63.09} & 46.18 & 57.43 & \underline{43.59} & 53.61 & \underline{48.37} & 41.97 & 65.63 & 24.45 & \underline{51.56} & 25.80 & 49.03 & \underline{64.62} & \underline{66.15} \\
    Gemini 3.6 Flash & \textbf{73.70} & \textbf{71.81} & \textbf{62.60} & \underline{67.33} & 42.05 & \textbf{87.95} & \textbf{59.48} & 47.73 & \underline{73.62} & 26.52 & \textbf{61.80} & 25.58 & \textbf{60.21} & \textbf{85.38} & 59.23 \\
    Qwen3.8-27B & \underline{54.74} & 48.99 & \underline{46.56} & 55.94 & 34.87 & 62.05 & 42.48 & 44.38 & 53.33 & 26.81 & 41.75 & 25.27 & 45.75 & 59.23 & 46.92 \\
    Qwen3-VL-8B-Instruct & 40.98 & 46.64 & 26.72 & 45.54 & 36.41 & 62.65 & 37.91 & 37.53 & 29.83 & 25.89 & 29.63 & 25.01 & 37.77 & 40.00 & \textbf{66.92} \\
    InternVL3.5-8B & 32.42 & 43.96 & 27.86 & 41.58 & 35.90 & 43.98 & 36.60 & 23.06 & 24.46 & 14.83 & 16.84 & 18.24 & 31.25 & 50.77 & 40.77 \\
    VideoLLaMA3-7B & 22.02 & 27.18 & 25.95 & 28.22 & 24.10 & 21.08 & 32.68 & 11.98 & 25.36 & 14.45 & 16.32 & 17.64 & 22.91 & 20.00 & 49.23 \\
    \midrule
    \rowcolor{gray!15}
    \multicolumn{16}{l}{\textit{\textbf{Text Summary as Memory}}} \\
    Seed-2.0-Lite & 49.54 & 48.32 & 38.17 & 62.38 & \underline{43.59} & 55.42 & \underline{48.37} & \underline{49.39} & 71.15 & 25.49 & 46.18 & \textbf{28.19} & 47.29 & -- & -- \\
    Gemini 3.6 Flash & 49.85 & 50.67 & 41.98 & \textbf{71.29} & \textbf{48.21} & \underline{84.94} & \textbf{59.48} & \textbf{54.20} & \textbf{75.68} & \textbf{28.82} & 51.01 & \underline{26.01} & \underline{53.54} & -- & -- \\
    Qwen3.8-27B & 45.26 & 38.59 & 39.31 & 52.97 & 34.36 & 59.04 & 46.41 & 41.59 & 56.63 & \underline{27.18} & 38.11 & 25.98 & 42.38 & -- & -- \\
    Qwen3-VL-8B-Instruct & 30.58 & 39.60 & 24.81 & 48.02 & 33.85 & 59.64 & 42.48 & 38.35 & 34.25 & 23.32 & 31.01 & 25.64 & 36.06 & -- & -- \\
    InternVL3.5-8B & 25.69 & 31.21 & 27.48 & 45.54 & 37.95 & 54.22 & 39.87 & 27.34 & 29.93 & 15.00 & 17.44 & 18.80 & 31.41 & -- & -- \\
    VideoLLaMA3-7B & 19.88 & 21.14 & 28.24 & 35.64 & 31.28 & 25.30 & 32.03 & 11.66 & 28.35 & 15.86 & 13.03 & 17.11 & 23.78 & -- & -- \\
    \specialrule{0.6pt}{0.4ex}{0pt}
    \specialrule{0.6pt}{0.4ex}{0.6ex}
    \rowcolor{gray!15}
    \multicolumn{16}{l}{\textit{\textbf{KV Cache}}} \\
    HERMES & \underline{36.09} & 34.23 & 32.06 & 39.60 & 33.85 & \underline{69.88} & 24.84 & \underline{35.82} & \underline{24.64} & \underline{17.39} & 19.08 & 18.34 & \underline{33.07} & 7.69 & \textbf{90.77} \\
    ReKV & 27.52 & 34.90 & 31.30 & 30.69 & 34.36 & 39.76 & 26.80 & 22.72 & 15.26 & 17.00 & 19.75 & \underline{19.28} & 27.65 & 10.77 & 63.85 \\
    \midrule
    \rowcolor{gray!15}
    \multicolumn{16}{l}{\textit{\textbf{Visual Tokens / Features}}} \\
    FLUXMem & 32.42 & \underline{41.95} & \underline{33.59} & 26.73 & 20.51 & 34.34 & 28.76 & 29.25 & 23.13 & 17.17 & 20.04 & 18.55 & 28.40 & 15.38 & 71.54 \\
    Flash-VStream & 29.97 & 38.59 & 27.10 & 31.19 & 33.85 & 32.53 & 25.49 & 5.32 & 17.70 & 10.82 & 13.07 & 10.15 & 24.69 & 7.69 & 52.31 \\
    \midrule
    \rowcolor{gray!15}
    \multicolumn{16}{l}{\textit{\textbf{Event Tree}}} \\
    StreamForest & 34.86 & \textbf{42.62} & \textbf{35.88} & \underline{45.05} & \textbf{42.05} & 56.02 & \underline{31.37} & 8.42 & 10.79 & 4.01 & 6.90 & 2.30 & 29.30 & 15.38 & 45.38 \\
    OASIS & \textbf{36.39} & 40.60 & 29.39 & \textbf{57.43} & \underline{36.41} & \textbf{74.70} & \textbf{43.14} & \textbf{52.93} & \textbf{52.98} & \textbf{22.07} & \textbf{28.90} & \textbf{23.06} & \textbf{41.46} & \textbf{23.08} & \underline{86.15} \\
    \midrule
    \rowcolor{gray!15}
    \multicolumn{16}{l}{\textit{\textbf{Parametric Memory}}} \\
    Video-Salmon-S & 17.43 & 15.10 & 12.98 & 17.33 & 20.00 & 31.93 & 22.22 & 22.61 & 23.22 & 15.16 & 14.65 & 14.98 & 18.72 & \underline{16.15} & 25.38 \\
    \midrule
    \rowcolor{gray!15}
    \multicolumn{16}{l}{\textit{\textbf{Reasoning Thoughts}}} \\
    VST & 20.80 & 23.15 & 17.94 & 10.40 & 29.23 & 16.27 & 11.11 & 5.00 & 17.65 & 16.62 & \underline{20.48} & 16.42 & 17.54 & 10.77 & 55.38 \\
    \bottomrule
  \end{tabular}%
  }
\end{table}

To simulate such intermittent real-world use, we introduce \textbf{APM-Bench}, which organizes egocentric experience into multi-session life trajectories, as shown in Figure~\ref{fig:teaser}. Sessions within a trajectory contain related activities and preserve their temporal order and time gaps. Each session includes fine-grained annotations of evidence time intervals, query times, reference proactive responses, etc. This organization naturally reduces the total video duration relative to a complete life log and enables us to compare memory effectiveness, storage costs, and response latency. APM-Bench evaluates three capabilities through 12 tasks: \emph{Cross-session Understanding} measures how well models use memory to answer questions about past sessions; \emph{Real-time Perception} assesses models' ability to understand the current visual scene and how memory affects this ability; and \emph{Adaptive Response} evaluates whether models provide appropriate help when needed and remain silent otherwise, with the help of memory.
\setlength{\intextsep}{\arxivOriginalIntextSep}

APM-Bench highlights three challenges in building effective persistent memory for real-world assistants. First, \textbf{\emph{what to store}}: to make past experience available in later sessions, the simplest strategy is to retain raw video, while compact persistent memory must selectively preserve information and represent it in forms such as visual tokens, structured events, or model parameters. Second, \textbf{\emph{when to use}}: persistent memory is not necessary for every interaction, and irrelevant historical information can interfere with current perception~\citep{arxiv260402317,arxiv260616353}. Third, \textbf{\emph{efficiency}}: the latency and storage costs introduced by persistent memory must remain manageable. Moreover, persistent memory cannot retain an unbounded visual history under finite storage, so assistants should acknowledge when relevant evidence is unavailable rather than fabricating an answer.

 We evaluate general video models by replaying stored videos at query time (\emph{video as memory}) or session summaries supplied with the query (\emph{text as memory}), alongside specialized memory systems. Video as memory preserves visual details but requires more storage, increases latency, and often weakens real-time perception; specialized systems vary widely in efficiency, yet most deliver weak service quality, as illustrated in Figure~\ref{fig:tradeoff}. Tests with unavailable historical evidence further show that even the strongest general video model struggles to acknowledge insufficient evidence.

Our contributions are as follows:
\begin{itemize}
    \item We introduce APM-Bench, with 2,719 human-refined candidates spanning objective and open-ended questions, averaging 69 minutes of video per trajectory.
    \item APM-Bench highlights challenges in making persistent memory practical and jointly evaluates cross-session understanding, real-time perception, and adaptive response within each trajectory, with controlled tests of responses to unavailable historical evidence.
    \item We systematically analyze memory representations and systems across task performance, storage, and latency, revealing the strengths and limitations of existing systems.
\end{itemize}
We hope to encourage research on streaming persistent memory that jointly considers utility, latency, and storage, and to support more usable memory systems for real-world assistants.

\section{Related Work}

\paragraph{Streaming Video Benchmarks.}

Streaming video benchmarks require models to respond as video arrives, using only what they have seen. StreamingBench~\citep{arxiv241103628} and OVO-Bench~\citep{arxiv250105510} evaluate online understanding; RTV-Bench~\citep{arxiv250502064} examines continuous perception and reasoning. For interaction, RIVER~\citep{arxiv260303985} and EgoSAT~\citep{arxiv260624422} combine retrospective, current, and prospective tasks; PhoStream~\citep{arxiv260122575} studies mobile scenarios; StreamArena~\citep{arxiv260805703} studies hour-scale interaction. Memory and proactive service are also evaluated: EgoStream~\citep{arxiv260531557} tests episodic recall across horizons, and EgoServe~\citep{arxiv260711523} tests long-term proactive service. ESTP-Bench~\citep{arxiv251014560}, EgoPro-Bench~\citep{arxiv260507299}, and OmniPro~\citep{arxiv260518577} further assess proactive response timing and content. These benchmarks move streaming evaluation toward real assistance, but mostly use single continuous videos, as shown in Table~\ref{tab:benchmark-comparison}. APM-Bench asks how retained memory supports later interactions across temporally separated sessions, and at what storage and latency cost.

\paragraph{Streaming Memory Systems.}

Streaming video memory methods have been extensively studied to determine what to retain from a growing visual stream. Visual representation methods compress features or tokens before passing them to the model \citep{arxiv260507897,arxiv260517921} or incrementally update stored representations as new frames arrive \citep{arxiv260827881,arxiv260904131}. Structured memories organize history around events \citep{streamforest2025,oasis2026} or objects and their state changes \citep{arxiv260728312}. KV-cache approaches retrieve or compress past states \citep{rekv2025,arxiv251018269} and organize them hierarchically \citep{hermes2026}. Text-based approaches preserve reasoning traces \citep{arxiv260311896,arxiv260312938} or structured summaries \citep{arxiv260830294} for later use. Parametric memory systems update model parameters during streaming \citep{arxiv251011129,arxiv260813416}. Yet most approaches manage context within a single continuous video. In real-world use, models need memory that can persist through interruptions, and remain useful when interactions resume. EgoMemo~\citep{arxiv260711523}, GROVE~\citep{arxiv260802392}, and StreamMind~\citep{arxiv260805703} construct persistent memory online, but retrieval and reasoning over that memory introduce substantial response latency, weakening time-sensitive adaptive responses. Questions remain about when to use memory, whether persistent state can be stored affordably, and how to respond when required evidence is unavailable.

\section{APM-Bench}
APM-Bench organizes egocentric video streams into multi-session trajectories of related activities, preserving continuous temporal order within sessions and realistic time gaps between them. The benchmark systematically evaluates three complementary capabilities: (1) \emph{Cross-session Understanding} assesses whether persistent memory reliably retains essential information across completed sessions; (2) \emph{Real-time Perception} evaluates streaming perception in the current scene, while examining whether incorporating persistent memory impacts real-time perception performance; and (3) \emph{Adaptive Response} tests whether the model can bridge current scenes with persistent memory to deliver timely, proactive responses. Across these capabilities, we introduce 12 task types categorized into intra-session or inter-session settings based on evidence location, where tasks in the first two families are formulated as multiple-choice questions, while all tasks in Adaptive Response are structured as open-ended questions. Furthermore, a dedicated evaluation set is constructed to examine whether models recognize when required evidence is unavailable rather than fabricate an answer.

\subsection{Benchmark Construction}

\textbf{Data Source.}
APM-Bench builds on two egocentric datasets: EgoLife~\citep{egolife2025}, with multi-day recordings, transcripts, and timestamped captions, and HD-EPIC~\citep{hdepic2025}, with fine-grained action and object annotations for structured kitchen procedures.

\textbf{Construction Pipeline.}
As illustrated in Figure~\ref{fig:benchmark-construction}, APM-Bench is constructed in two stages. In \emph{Stage~1}, we organize EgoLife and HD-EPIC videos into activity-related multi-session trajectories with real-world timestamps. In \emph{Stage~2}, we generate candidates for the three capability families and filter and refine them through automated and human review, yielding 2{,}719 candidates. On 300 sampled questions, two annotators reach a Cohen's kappa of $0.868$~\citep{cohen1960coefficient}.

\textbf{Evidence Availability-Aware Evaluation Set.}
Finite storage and compression prevent persistent memory from retaining all visual history. We therefore curate 260 Cross-session Understanding questions to test whether models recognize unavailable evidence. Models access only the two most recent completed sessions before the query: 130 questions have all required evidence within this history, while the other 130 require evidence outside it. Each question includes a coarse time span and a fifth option indicating insufficient available evidence. This setting tests whether models answer when evidence is accessible and acknowledge when it is not.

\subsection{Detail of APM-Bench}
As illustrated in Figure~\ref{fig:task-examples}, we characterize six streaming formulations of queries, evidence, spanning intra-session and inter-session settings based on evidence location.

\textbf{Cross-session Understanding} requires evidence from previous sessions. \emph{Episodic Recall} (ER) recalls or summarizes events and activities from earlier sessions. \emph{Entity State Tracking} (EST) tracks the states or locations of objects and other entities over time. \emph{Temporal Reasoning} (TR) compares multiple historical moments to infer event orderings and temporal changes. Tasks in this family are formulated as multiple-choice questions and span single- and multi-evidence temporal grounding.

\textbf{Real-time Perception} evaluates understanding of the current visual scene within the ongoing session. \emph{Action Recognition} (ACR) identifies actions performed by the wearer or nearby individuals. \emph{Counting} (CT) counts instances of specified entities. \emph{Optical Character Recognition} (OCR) reads visible text, labels, or screen content. \emph{Spatial Understanding} (STU) determines spatial relationships among specific entities. Tasks in this family are also formulated as multiple-choice questions.

\textbf{Adaptive Response} evaluates whether the model delivers timely assistance when trigger conditions are met and remains silent otherwise. \emph{Evidence-Ready Answering} (ERA) releases a multiple-choice question before its causal evidence appears; the model must remain \texttt{SILENT} until the evidence is sufficient, then \texttt{INTERVENE} with the selected option and rationale. \emph{Registered-Condition Response} (RCR) requires the model to detect whether the ongoing scene fulfills a reminder condition registered earlier. \emph{Memory-Grounded Proactive Assistance} (MPA) leverages past experience without explicit user instructions to offer proactive guidance during related activities, requiring models to connect historical memory with current actions. \emph{Proactive Reminder} (PRM) triggers a reminder registered in a prior session when conditions arise in a later session. \emph{Task Progress Guidance} (TPG) tracks progress across long-running tasks and delivers task-relevant assistance upon resumption. ERA and RCR are \emph{intra-session} tasks confined to the current session, whereas MPA, PRM, and TPG are \emph{inter-session} tasks that bridge current visual events with persistent historical memory. Except for ERA, the other tasks require outputting the decision, rationale, and response. All adaptive tasks are evaluated in an open-ended format via LLM-as-a-judge.

As shown in Figure~\ref{fig:benchmark-distributions}, each trajectory contains 69 minutes of video on average, with individual sessions averaging 13 minutes, while its real-world span can extend across hours or days due to inter-session gaps. This separation between video duration and elapsed real-world time reflects the intermittent interactions that persistent memory must support, with more statistics in Appendix~\ref{app:statistics}.

\subsection{Evaluation Protocol and Metrics}
\label{sec:evaluation-protocol}

\paragraph{Online Inference Protocol}
During inference, models access prior-session persistent memory and the current session's causal video prefix ending at the probe or query timestamp, strictly adhering to causal constraints. For Adaptive Response, each candidate contains probes at precise timestamps labeled as \texttt{SILENT} or \texttt{INTERVENE}, treating each probe as an independent runtime instance (3{,}768 in total). Specifically, ERA releases its multiple-choice question at session start without repeating it at subsequent probes; for MPA and TPG, probes provide brief task definitions to guide decisions; RCR and PRM inject textual reminder instructions at registration timestamps. Conversely, candidates in Cross-session Understanding (887) and Real-time Perception (716) each form a single runtime instance evaluated at the query timestamp, requiring the model to directly output option choices.

\paragraph{Metrics}

Table~\ref{tab:evaluation-metrics} summarizes the evaluation metrics. For \emph{Adaptive Response}, we propose a Gated LLM-Judge Score across all five tasks. A response passes the gate if it correctly decides to \texttt{INTERVENE} or remain \texttt{SILENT}; for positive ERA probes, the predicted MCQA option must additionally match the ground truth. Probes failing the gate receive $g_r=0$. For gate-passing responses, DeepSeek-V4-Flash evaluates the rationale against the reference response and assigns a score $g_r\in\{1,\ldots,5\}$.For candidate $c$, let $\mathcal{P}_c$ and $\mathcal{N}_c$ denote its positive and negative probe sets, with mean scores $\bar{g}_c^+ = \frac{1}{|\mathcal{P}_c|}\sum_{r\in\mathcal{P}_c} g_r$ and $\bar{g}_c^- = \frac{1}{|\mathcal{N}_c|}\sum_{r\in\mathcal{N}_c} g_r$, respectively. The candidate-level score $s_c$ and the task-level score over the candidate set $C_t$ are defined as:

\begin{equation}
\mathrm{GatedLLMJudge}_t = \frac{20}{|C_t|}\sum_{c\in C_t} s_c, \quad
s_c =
\left\{
\begin{array}{ll}
\frac{1}{2}(\bar{g}_c^+ + \bar{g}_c^-), & \mathcal{P}_c \neq \emptyset \land \mathcal{N}_c \neq \emptyset, \\
\bar{g}_c^{\star}, & \text{otherwise}.
\end{array}
\right.
\label{eq:adaptive-gated}
\end{equation}
where $\bar{g}_c^{\star}$ denotes the mean score over the sole available probe set, either $\mathcal{P}_c$ or $\mathcal{N}_c$. The factor of 20 converts candidate scores to a 0--100 task score. See Appendix~\ref{app:judge} for the judging rubric.

\emph{Storage cost} quantifies the persistent state retained across completed prior sessions, strictly excluding the ongoing session. For each trajectory $\tau$, we capture the peak prior-session storage footprint $B_\tau^{\text{peak}}$ normalized by the cumulative prior-session duration $D_\tau^{\text{peak}}$ in hours:
\begin{equation}
\text{Storage/hr} = \frac{1}{|\mathcal{T}|}\sum_{\tau \in \mathcal{T}} \frac{B_\tau^{\text{peak}}}{D_\tau^{\text{peak}}}.
\label{eq:storage-cost}
\end{equation}

\section{Experiment}

\subsection{Experiment Setup}
\label{sec:experiment-setup}

\paragraph{General Large Video Models}
We evaluate a no-memory baseline and two memory conditions for general video models. \emph{SimpleStream} uses Qwen3-VL-8B-Instruct~\citep{qwen3vl2025} with only the four most recent frames as the no-memory baseline. \emph{Raw Video as Memory} stores all prior-session videos and replays them at query time. \emph{Text Summary as Memory} generates a summary at the end of each session and provides prior-session summaries together with the current causal video prefix at query time. All models use 1\,FPS video input. Proprietary models and Qwen-series models uniformly sample up to 1{,}024 frames, while InternVL3.5-8B~\citep{internvl352025} and VideoLLaMA3-7B~\citep{videollama32025} sample up to 128 frames.
\paragraph{Specialized Streaming Memory Systems}
We evaluate eight methods across five memory representations: \emph{KV Cache}, \emph{Visual Tokens/Features}, \emph{Event Tree}, \emph{Parametric Memory}, and \emph{Reasoning Thoughts}. All use official settings and process prior sessions sequentially. For methods supporting streaming input, latency is measured from the last required memory update to the first output token; otherwise, from processing the causal video prefix to the first output token. As most methods target single continuous videos without persistent-state export, Table~\ref{tab:efficiency-analysis} estimates storage from the memory state maintained during inference. Implementation details are in Appendix~\ref{app:compute} and \ref{app:backbones}.

\subsection{Main Results}

\begin{insightbox}{APMInsightBlue}{What to store is a key design choice for persistent memory.}
Storing raw video is the simplest strategy and achieves the strongest performance for most general models, but incurs substantial storage and latency costs. Event-structured methods lead among specialized systems, showing their potential as an alternative to raw-video replay.
\end{insightbox}
Tables~\ref{tab:efficiency-analysis} and~\ref{tab:main-results} show that richer retained information benefits Cross-session Understanding. Without prior-session memory, \emph{SimpleStream} performs substantially worse on cross-session tasks. \emph{Raw Video as Memory} preserves the richest visual evidence and achieves the strongest cross-session performance, whereas \emph{Text Summary as Memory} loses visual details during summarization and degrades cross-session understanding. Specialized systems adopt different memory representations, including KV caches, visual tokens or features, event structures, parametric memory, and reasoning thoughts. Among them, \textbf{\emph{event-structured}} methods achieve the strongest utility, suggesting that organizing history around events is a promising alternative to raw-video replay. \textbf{\emph{Parametric memory}} is also appealing because its state does not grow with video length, although the evaluated method still suffers from limited utility and nontrivial latency. By organizing intermittent interactions into sessions, APM-Bench naturally defines memory-storage boundaries and enables clearer comparison across memory representations, as further analyzed in Appendix~\ref{app:organization}.

\begin{insightbox}{APMInsightGreen}{Real-world deployment requires efficient persistent memory.}
Existing memory systems rarely achieve strong utility, low latency, and low storage cost simultaneously. Practical persistent memory therefore requires jointly optimizing all three dimensions rather than improving any one of them in isolation.
\end{insightbox}
The activity-related trajectory design preserves long real-world spans while reducing the amount of video history, making storage and latency easier to measure across methods. Table~\ref{tab:efficiency-analysis} shows that storage and latency are closely coupled with utility. \emph{Raw-video memory} achieves strong performance but requires GiB-scale storage and incurs high query-time overhead, while \emph{text summaries} reduce storage to the KiB scale at the cost of cross-session performance. Specialized memory systems improve efficiency in different ways, but compact or fast methods often sacrifice utility, while stronger systems can require substantially more storage or response time. These results reveal a clear utility--latency--storage trade-off, highlighting the need to jointly consider all three dimensions when designing persistent memory for real-world streaming assistants.

\begin{insightbox}{APMInsightPurple}{Persistent memory should be used selectively.}
Not every interaction benefits from historical information, especially for real-time perception. Historical information should be injected only when relevant to the ongoing interaction.
\end{insightbox}
Real-time Perception depends on the current visual scene and often does not require prior-session history. As shown in Table~\ref{tab:main-results}, replacing raw-video history with text summaries improves Real-time Perception for most general video models, suggesting that excessive or irrelevant history can interfere with current-scene understanding. This reveals a tension between maintaining rich historical context and accurate real-time perception: information useful for later recall may be unnecessary or distracting in the current interaction.Retaining more history is therefore insufficient; the system must also determine whether that history is useful now and what should be exposed to the model. Persistent memory should therefore be used selectively, injecting historical information only when it is relevant to the ongoing interaction.

\begin{insightbox}{APMInsightYellow}{Reliable deployment requires awareness of evidence availability.}
Answering with available evidence and recognizing missing evidence are distinct capabilities. Persistent memory systems should make the coverage of retained history explicit so that assistants can identify when required evidence is missing.
\end{insightbox}
On the \emph{Evidence Availability-Aware} evaluation, Gemini~3.6~Flash performs strongly when the required evidence remains accessible but is less reliable at recognizing when it is unavailable. SimpleStream shows the opposite pattern: with only the four most recent frames, it strongly favors the insufficient-evidence option but rarely answers correctly when historical evidence is available. These contrasting cases show that successful recall does not imply reliable awareness of what evidence remains accessible. One practical direction is to make the coverage of retained history explicit, enabling the assistant to recognize when required evidence is unavailable rather than fabricate an answer. Appendix~\ref{app:evidence-availability} further compares the original 4-way MCQA accuracy of questions used to construct this set with their 5-way accuracy when the required evidence remains available.

\subsection{Why Adaptive Response Remains Difficult}

Adaptive Response remains the most challenging capability even for the strongest general video models. The task-level results in Table~\ref{tab:main-results} show a clear distinction between assistance driven by explicit registrations and fully autonomous proactive assistance. RCR and PRM provide the model with a previously registered condition or reminder, giving it a concrete target to monitor in the current scene. Models perform substantially better under these paradigms than on MPA and TPG, where no explicit trigger specifies which historical experience should become relevant.

MPA and TPG require models to connect relevant past experience to the current scene, decide whether to intervene, and provide useful assistance. Richer history alone does not solve this: even with Raw Video as Memory, the strongest general model remains weak on these tasks despite much stronger Cross-session Understanding. As further shown in Appendix~\ref{app:organization}, even when given only the necessary historical evidence, models still struggle on \emph{Adaptive Response}, indicating that relating past evidence to the current scene and turning it into useful assistance remains difficult. Further analyses in Appendix~\ref{app:decision} and Appendix~\ref{app:conditional} examine decision accuracy and response quality after correct decisions, respectively.

\section{Conclusion}
We introduce APM-Bench, which organizes egocentric experience into activity-related multi-session trajectories to simulate intermittent real-world use and evaluate persistent memory in streaming video models. A useful persistent memory must be storable and reusable across sessions, while carefully balancing what to store, when to use it, and efficiency.
Our experiments reveal clear trade-offs. Raw video preserves the richest visual evidence and achieves the strongest cross-session performance, but incurs substantial storage and latency costs. Compact representations reduce these costs but may lose details needed later; among specialized systems, event-structured memory shows the strongest utility. Excessive history can impair real-time perception, motivating selective memory use. Moreover, strong recall ability does not guarantee awareness when required evidence is unavailable, and adaptive response remains challenging even with rich historical context. Overall, current methods still struggle to simultaneously achieve reliable long-term recall, selective memory use, low storage and latency overhead, and effective proactive assistance across sessions.

\bibliography{iclr2027_conference}
\bibliographystyle{iclr2027_conference}

\clearpage
\appendix
\section*{Appendix Contents}
\newcommand{\appcontentssection}[2]{\noindent\hyperref[#1]{\textbf{\ref*{#1}\quad #2}}\dotfill\pageref*{#1}\par\smallskip}
\newcommand{\appcontentssubsection}[2]{\noindent\hspace*{1.5em}\hyperref[#1]{\ref*{#1}\quad #2}\dotfill\pageref*{#1}\par}
\begingroup
\appcontentssection{app:implementation}{Implementation Details}
\appcontentssubsection{app:compute}{Compute}
\appcontentssubsection{app:backbones}{Backbones}
\appcontentssubsection{app:judge}{LLM-as-Judge rubric}
\appcontentssection{app:results}{Additional Results and Analysis}
\appcontentssubsection{app:decision}{Adaptive decision accuracy}
\appcontentssubsection{app:conditional}{Response quality after correct decisions}
\appcontentssubsection{app:latency}{Latency}
\appcontentssubsection{app:storage}{Persistent storage}
\appcontentssubsection{app:invalid}{Instruction following}
\appcontentssubsection{app:organization}{Effect of trajectory organization}
\appcontentssubsection{app:evidence-availability}{Evidence availability and answer accuracy}
\appcontentssubsection{app:profiles}{Task-level performance}
\appcontentssection{app:statistics}{Dataset Statistics}
\appcontentssubsection{app:activities}{Trajectory activities}
\appcontentssubsection{app:evidence}{Evidence composition}
\appcontentssubsection{app:session-distance}{Evidence distance measured in sessions}
\appcontentssubsection{app:span}{Real-world span and organized trajectory duration}
\appcontentssection{app:data}{Data Construction and Annotation Details}
\appcontentssubsection{app:dataset-construction}{Dataset construction}
\appcontentssubsection{app:verification}{Human verification}
\appcontentssubsection{app:candidates}{MCQA annotation format}
\appcontentssubsection{app:era}{ERA: evidence readiness}
\appcontentssubsection{app:rcr}{RCR: in-session conditional reminder}
\appcontentssubsection{app:mpa}{MPA: assistance from prior experience}
\appcontentssubsection{app:prm}{PRM: cross-session registered reminder}
\appcontentssubsection{app:tpg}{TPG: continuing-task guidance}
\appcontentssection{app:prompts}{Task Prompts}
\appcontentssubsection{app:system-prompt}{System prompt}
\appcontentssubsection{app:mcqa-prompt}{MCQA prompts}
\appcontentssubsection{app:adaptive-prompt}{Adaptive Response probes}
\appcontentssubsection{app:evidence-availability-prompt}{Evidence Availability-Aware prompt}
\appcontentssubsection{app:text-summary-prompt}{Text-summary memory prompt}
\endgroup

\clearpage
\raggedbottom
\section{Implementation Details}
\label{app:implementation}

\subsection{Compute}
\label{app:compute}
All experiments ran on H20 GPUs with the same host configuration, enabling a fair comparison of latency and storage cost across open-source video models and specialized memory systems.

\subsection{Backbones}
\label{app:backbones}
Table~\ref{tab:appendix-backbones} lists the backbone of each specialized system and SimpleStream.

\begin{table}[H]
\caption{Visual-language backbones of specialized memory systems and SimpleStream.}
\label{tab:appendix-backbones}
\centering\small
\begin{tabular}{@{}ll@{}}
\toprule
\textbf{System} & \textbf{Visual-language backbone} \\
\midrule
SimpleStream & Qwen3-VL-8B-Instruct \\
HERMES & Qwen2.5-VL-7B-Instruct \\
ReKV & LLaVA-OneVision-Qwen2-7B \\
FluxMem & Qwen2.5-VL-7B-Instruct \\
Flash-VStream & Qwen2-VL-7B \\
StreamForest & Qwen2-7B \\
OASIS & Qwen3-VL-8B-Instruct \\
Video-SALMONN S & Qwen3-VL-8B \\
VST & Qwen2.5-VL-7B-Instruct \\
\bottomrule
\end{tabular}
\end{table}

\subsection{LLM-as-Judge rubric}
\label{app:judge}
After the deterministic gate in \hyperref[sec:evaluation-protocol]{Section~\ref*{sec:evaluation-protocol}}, DeepSeek V4 Flash scores response quality using the rubric and task guidance below. The judge receives the causal cutoff, reference annotations, and model response for each probe.

\begin{promptbox}{DeepSeek V4 Flash: complete judge system rubric and output contract}
You are an impartial evaluator for a streaming video assistant benchmark.

The assistant's SILENT/INTERVENE decision has already passed deterministic correctness checks. Evaluate only the semantic quality of its explanation and, when it intervenes, its proactive response. Reference annotations describe accepted evidence and one acceptable response; do not require lexical overlap or identical wording. Do not reward verbosity. Do not infer facts outside the supplied annotations.

For a correct SILENT decision, judge whether the explanation identifies the decisive condition that is still missing and distinguishes this evaluation window from a valid response opportunity. For ERA, evaluate only whether the reason correctly explains whether the answer evidence is causally available; the MCQA answer itself has already passed deterministic checking.

Use this 1-5 scale:
5 = task-faithful, fully grounded, factually accurate, precise, and useful.
4 = core content is correct and useful, with only a minor omission or imprecision.
3 = basically correct but generic, incomplete, or uses only part of the important evidence.
2 = related but has a major grounding gap, factual issue, or omission that could mislead.
1 = the decision happens to be correct, but the explanation or response is substantially inconsistent or unusable.

Return exactly one JSON object with all and only the following fields:
{
  "score": 1,
  "criterion_checks": {
    "task_fidelity": "pass | partial | fail",
    "historical_or_registration_grounding": "pass | partial | fail | not_applicable",
    "current_trigger_or_silence_grounding": "pass | partial | fail",
    "factual_support": "pass | partial | fail",
    "usefulness": "pass | partial | fail | not_applicable"
  },
  "critical_issues": [],
  "justification": "Concise evidence-based explanation."
}

The value shown as 1 for score is an example; replace it with one integer from 1 through 5. For each criterion, return exactly one of the literal enum values separated by | above, not the whole displayed string. Include every criterion_checks key even when its value is not_applicable. Keep critical_issues short and use an empty array when there is no critical issue. Keep justification concise and evidence-based. Do not wrap the JSON in Markdown or add text before or after it.
\end{promptbox}

\begin{promptbox}{DeepSeek V4 Flash: task-specific guidance}
Task-specific guidance field (select the matching task):

ERA: Check whether the reason correctly explains that answer evidence is or is not causally available at this heartbeat.

RCR: Check fidelity to the in-session registration and whether the current observable condition justifies the registered response.

MPA: Check whether earlier experience materially improves assistance for the current scene and whether the response is actionable rather than generic.

PRM: Check fidelity to the earlier registration, whether its observable condition is satisfied, and whether the response preserves the obligation.

TPG: Check whether history and the current scene belong to the same continuing task and whether progress or next-step guidance is accurate.
\end{promptbox}

\section{Additional Results and Analysis}
\label{app:results}

\subsection{Adaptive decision accuracy}
\label{app:decision}
For task $t$, let $P_t$ be the fraction of positive probes with a correct \texttt{INTERVENE} decision and $N_t$ the fraction of negative probes with a correct \texttt{SILENT} decision. Decision balanced accuracy is $(P_t+N_t)/2$. The overall rate pools probes across the five Adaptive tasks before averaging the positive and negative rates. This measure tests whether the model acts at the right time; the main Gated Judge score also evaluates the answer or response under the \hyperref[sec:evaluation-protocol]{Section~\ref*{sec:evaluation-protocol}} protocol. ERA's positive decision rate alone does not require the MCQA option to be correct. Table~\ref{tab:appendix-decision} reports the results.

\begin{table}[H]
\caption{Adaptive decision accuracy (\%). Each task score averages the correct \texttt{INTERVENE} rate on positive probes and the correct \texttt{SILENT} rate on negative probes. Overall pools probes across the five tasks before reporting the positive rate ($P$), negative rate ($N$), and their balanced average ($BA$).}
\label{tab:appendix-decision}
\centering\setlength{\tabcolsep}{4pt}\renewcommand{\arraystretch}{1.08}
\resizebox{\textwidth}{!}{%
\begin{tabular}{@{}lcccccc@{}}
\toprule
\textbf{Model / Method} & \textbf{ERA} & \textbf{RCR} & \textbf{MPA} & \textbf{PRM} & \textbf{TPG} & \textbf{Overall $P/N/BA$} \\
\midrule
\rowcolor{gray!15}
\multicolumn{7}{l}{\textit{\textbf{w/o Memory}}} \\
SimpleStream & 65.89 & 50.81 & 54.53 & 50.86 & 50.39 & 19.86/92.81/56.34 \\
\specialrule{0.6pt}{0.4ex}{0pt}
\specialrule{0.6pt}{0.4ex}{0.6ex}
\rowcolor{gray!15}
\multicolumn{7}{l}{\textit{\textbf{Raw Video as Memory}}} \\
Seed-2.0-Lite & 62.10 & 73.63 & 53.94 & 69.56 & 58.15 & 80.66/46.65/63.66 \\
Gemini 3.6 Flash & 63.80 & 82.76 & 51.58 & 77.55 & 52.16 & 74.33/59.04/66.69 \\
Qwen3-VL-8B & 58.49 & 51.47 & 56.80 & 52.58 & 58.18 & 64.79/44.28/54.54 \\
Qwen3.8-27B & 68.44 & 67.62 & 52.37 & 59.97 & 52.62 & 53.69/71.01/62.35 \\
InternVL3.5-8B & 52.42 & 50.00 & 50.75 & 50.86 & 52.77 & 90.37/13.42/51.90 \\
VideoLLaMA3-7B & 37.62 & 51.10 & 51.69 & 50.31 & 51.12 & 84.65/9.33/46.99 \\
\midrule
\rowcolor{gray!15}
\multicolumn{7}{l}{\textit{\textbf{Text Summary as Memory}}} \\
Seed-2.0-Lite & 67.77 & 78.30 & 53.29 & 65.84 & 54.65 & 77.10/55.60/66.35 \\
Gemini 3.6 Flash & 70.39 & 84.21 & 54.56 & 70.53 & 51.56 & 71.55/66.88/69.22 \\
Qwen3-VL-8B & 60.27 & 53.41 & 58.21 & 57.15 & 60.53 & 74.33/41.26/57.79 \\
Qwen3.8-27B & 68.07 & 67.53 & 55.33 & 62.52 & 53.40 & 64.53/61.26/62.89 \\
InternVL3.5-8B & 53.19 & 49.69 & 49.80 & 51.49 & 50.94 & 80.66/24.51/52.59 \\
VideoLLaMA3-7B & 40.71 & 50.00 & 50.28 & 51.40 & 50.88 & 87.51/7.72/47.62 \\
\specialrule{0.6pt}{0.4ex}{0pt}
\specialrule{0.6pt}{0.4ex}{0.6ex}
\rowcolor{gray!15}
\multicolumn{7}{l}{\textit{\textbf{KV Cache}}} \\
HERMES & 50.88 & 54.26 & 53.49 & 52.75 & 52.34 & 58.02/51.97/55.00 \\
ReKV & 52.80 & 35.67 & 47.29 & 42.60 & 49.77 & 52.99/34.26/43.63 \\
\midrule
\rowcolor{gray!15}
\multicolumn{7}{l}{\textit{\textbf{Visual Tokens / Features}}} \\
FluxMem & 55.54 & 52.29 & 53.44 & 46.86 & 51.97 & 74.67/32.35/53.51 \\
Flash-VStream & 36.47 & 50.28 & 49.96 & 49.77 & 48.82 & 24.11/61.38/42.74 \\
\midrule
\rowcolor{gray!15}
\multicolumn{7}{l}{\textit{\textbf{Event Tree}}} \\
StreamForest & 40.46 & 21.30 & 16.27 & 22.62 & 11.32 & 23.33/27.72/25.53 \\
OASIS & 72.87 & 69.28 & 57.90 & 54.69 & 56.97 & 74.15/60.84/67.50 \\
\midrule
\rowcolor{gray!15}
\multicolumn{7}{l}{\textit{\textbf{Parametric Memory}}} \\
Video-SALMONN S & 53.22 & 53.96 & 50.75 & 46.07 & 52.64 & 68.78/34.34/51.56 \\
\midrule
\rowcolor{gray!15}
\multicolumn{7}{l}{\textit{\textbf{Reasoning Thoughts}}} \\
VST & 11.45 & 46.17 & 51.42 & 53.61 & 48.77 & 61.49/9.06/35.28 \\
\bottomrule
\end{tabular}}
\end{table}

\subsection{Response quality after correct decisions}
\label{app:conditional}
After a probe passes the \hyperref[sec:evaluation-protocol]{Section~\ref*{sec:evaluation-protocol}} gate, we measure the quality of the answer or proactive response when the model speaks, and its reason for remaining silent otherwise. For each candidate, we average judged probes within each polarity, weight the available polarities equally, and then average over candidates with at least one judged probe. The judge's 1--5 score is converted to a 0--100 scale. Let $A_c$ be the set of polarities with a judged, gate-passing probe for candidate $c$. Then
\begin{equation}
\mathrm{ResponseQuality}_t=\frac{100}{5|C'_t|}\sum_{c\in C'_t}\frac{1}{|A_c|}\sum_{a\in A_c}\bar{j}_{c,a},
\label{eq:conditional-judge}
\end{equation}
where $C'_t$ contains candidates with $A_c\neq\emptyset$ and $\bar{j}_{c,a}$ is the mean judge score for candidate $c$ and polarity $a$. Table~\ref{tab:appendix-conditional} reports these scores separately from the main Gated Judge result. Its superscripts give the balanced gate pass rate: we compute the fraction of positive and negative probes passing the deterministic gate separately, then average the two fractions. For ERA, a positive probe passes only if the model intervenes and selects the correct MCQA option.

\begin{table}[H]
\caption{Response quality after the \hyperref[sec:evaluation-protocol]{Section~\ref*{sec:evaluation-protocol}} gate (\%). Judged probes are averaged within each available polarity, then by candidate, using Eq.~\ref{eq:conditional-judge}. Superscripts give the balanced gate pass rate for each task: the mean of the positive and negative probe pass rates. Positive ERA probes also require the correct MCQA answer. Average is the mean of the five task scores.}
\label{tab:appendix-conditional}
\centering\setlength{\tabcolsep}{4pt}\renewcommand{\arraystretch}{1.08}
\resizebox{\textwidth}{!}{%
\begin{tabular}{@{}lcccccc@{}}
\toprule
\textbf{Model / Method} & \textbf{ERA} & \textbf{RCR} & \textbf{MPA} & \textbf{PRM} & \textbf{TPG} & \textbf{Average} \\
\midrule
\rowcolor{gray!15}
\multicolumn{7}{l}{\textit{\textbf{w/o Memory}}} \\
SimpleStream & 84.45\textsuperscript{59.1\%} & 60.18\textsuperscript{50.8\%} & 49.04\textsuperscript{54.5\%} & 63.81\textsuperscript{50.9\%} & 48.11\textsuperscript{50.4\%} & 61.12 \\
\specialrule{0.6pt}{0.4ex}{0pt}
\specialrule{0.6pt}{0.4ex}{0.6ex}
\rowcolor{gray!15}
\multicolumn{7}{l}{\textit{\textbf{Raw Video as Memory}}} \\
Seed-2.0-Lite & 92.91\textsuperscript{45.4\%} & 88.52\textsuperscript{73.6\%} & 46.45\textsuperscript{53.9\%} & 72.52\textsuperscript{69.6\%} & 44.73\textsuperscript{58.2\%} & 69.02 \\
Gemini 3.6 Flash & 88.33\textsuperscript{54.2\%} & 88.84\textsuperscript{82.8\%} & 52.58\textsuperscript{51.6\%} & 78.64\textsuperscript{77.6\%} & 52.31\textsuperscript{52.2\%} & 72.14 \\
Qwen3-VL-8B & 78.80\textsuperscript{47.6\%} & 57.46\textsuperscript{51.5\%} & 46.47\textsuperscript{56.8\%} & 54.05\textsuperscript{52.6\%} & 43.84\textsuperscript{58.2\%} & 56.12 \\
Qwen3.8-27B & 83.84\textsuperscript{52.3\%} & 77.35\textsuperscript{67.6\%} & 52.23\textsuperscript{52.4\%} & 66.78\textsuperscript{60.0\%} & 50.64\textsuperscript{52.6\%} & 66.17 \\
InternVL3.5-8B & 83.28\textsuperscript{27.9\%} & 48.92\textsuperscript{50.0\%} & 29.04\textsuperscript{50.7\%} & 33.00\textsuperscript{50.9\%} & 31.98\textsuperscript{52.8\%} & 45.24 \\
VideoLLaMA3-7B & 72.50\textsuperscript{16.8\%} & 50.13\textsuperscript{51.1\%} & 27.74\textsuperscript{51.7\%} & 32.01\textsuperscript{50.3\%} & 32.12\textsuperscript{51.1\%} & 42.90 \\
\midrule
\rowcolor{gray!15}
\multicolumn{7}{l}{\textit{\textbf{Text Summary as Memory}}} \\
Seed-2.0-Lite & 94.22\textsuperscript{52.6\%} & 90.27\textsuperscript{78.3\%} & 48.24\textsuperscript{53.3\%} & 69.30\textsuperscript{65.8\%} & 53.07\textsuperscript{54.7\%} & 71.02 \\
Gemini 3.6 Flash & 86.59\textsuperscript{62.9\%} & 89.51\textsuperscript{84.2\%} & 52.85\textsuperscript{54.6\%} & 71.70\textsuperscript{70.5\%} & 54.83\textsuperscript{51.6\%} & 71.10 \\
Qwen3-VL-8B & 79.77\textsuperscript{47.7\%} & 63.83\textsuperscript{53.4\%} & 41.88\textsuperscript{58.2\%} & 53.41\textsuperscript{57.2\%} & 40.74\textsuperscript{60.5\%} & 55.93 \\
Qwen3.8-27B & 84.84\textsuperscript{48.9\%} & 82.22\textsuperscript{67.5\%} & 51.01\textsuperscript{55.3\%} & 59.32\textsuperscript{62.5\%} & 51.37\textsuperscript{53.4\%} & 65.75 \\
InternVL3.5-8B & 78.86\textsuperscript{34.3\%} & 60.41\textsuperscript{49.7\%} & 29.83\textsuperscript{49.8\%} & 33.58\textsuperscript{51.5\%} & 33.51\textsuperscript{50.9\%} & 47.24 \\
VideoLLaMA3-7B & 72.66\textsuperscript{16.3\%} & 56.95\textsuperscript{50.0\%} & 31.64\textsuperscript{50.3\%} & 25.22\textsuperscript{51.4\%} & 30.64\textsuperscript{50.9\%} & 43.42 \\
\specialrule{0.6pt}{0.4ex}{0pt}
\specialrule{0.6pt}{0.4ex}{0.6ex}
\rowcolor{gray!15}
\multicolumn{7}{l}{\textit{\textbf{KV Cache}}} \\
HERMES & 71.16\textsuperscript{49.9\%} & 44.67\textsuperscript{54.3\%} & 33.42\textsuperscript{53.5\%} & 35.90\textsuperscript{52.8\%} & 33.80\textsuperscript{52.3\%} & 43.79 \\
ReKV & 68.51\textsuperscript{33.5\%} & 42.82\textsuperscript{35.7\%} & 37.31\textsuperscript{47.3\%} & 47.95\textsuperscript{42.6\%} & 40.56\textsuperscript{49.8\%} & 47.43 \\
\midrule
\rowcolor{gray!15}
\multicolumn{7}{l}{\textit{\textbf{Visual Tokens / Features}}} \\
FluxMem & 72.50\textsuperscript{39.8\%} & 44.14\textsuperscript{52.3\%} & 31.61\textsuperscript{53.4\%} & 43.42\textsuperscript{46.9\%} & 34.53\textsuperscript{52.0\%} & 45.24 \\
Flash-VStream & 42.08\textsuperscript{12.9\%} & 34.88\textsuperscript{50.3\%} & 21.48\textsuperscript{50.0\%} & 26.23\textsuperscript{49.8\%} & 22.38\textsuperscript{48.8\%} & 29.41 \\
\midrule
\rowcolor{gray!15}
\multicolumn{7}{l}{\textit{\textbf{Event Tree}}} \\
StreamForest & 32.62\textsuperscript{25.3\%} & 51.85\textsuperscript{21.3\%} & 24.38\textsuperscript{16.3\%} & 30.83\textsuperscript{22.6\%} & 21.76\textsuperscript{11.3\%} & 32.29 \\
OASIS & 85.22\textsuperscript{61.7\%} & 76.11\textsuperscript{69.3\%} & 39.53\textsuperscript{57.9\%} & 51.39\textsuperscript{54.7\%} & 40.77\textsuperscript{57.0\%} & 58.61 \\
\midrule
\rowcolor{gray!15}
\multicolumn{7}{l}{\textit{\textbf{Parametric Memory}}} \\
Video-SALMONN S & 78.75\textsuperscript{28.8\%} & 41.72\textsuperscript{54.0\%} & 32.30\textsuperscript{50.7\%} & 31.31\textsuperscript{46.1\%} & 30.88\textsuperscript{52.6\%} & 42.99 \\
\midrule
\rowcolor{gray!15}
\multicolumn{7}{l}{\textit{\textbf{Reasoning Thoughts}}} \\
VST & 72.24\textsuperscript{6.7\%} & 38.23\textsuperscript{46.2\%} & 32.72\textsuperscript{51.4\%} & 38.17\textsuperscript{53.6\%} & 32.20\textsuperscript{48.8\%} & 42.71 \\
\bottomrule
\end{tabular}}
\end{table}

\subsection{Latency}
\label{app:latency}
For local model runs, query-to-first-token time includes memory preparation, visual processing, and the time until the first generated token, under the \hyperref[sec:experiment-setup]{Section~\ref*{sec:experiment-setup}} input settings. We additionally report an offline replay real-time factor, defined as the processing time from the beginning of a causal video prefix to the first output token divided by that prefix's video duration; video decoding is excluded from the processing time. Values below one indicate that processing can keep pace with the video clock under this replay measurement. Table~\ref{tab:appendix-rtf} gives per-method means for the eight specialized systems. For proprietary APIs, we measure end-to-end client latency from request submission until the complete response; the full-response means are in Table~\ref{tab:appendix-api-latency}.

\begin{table}[H]
\caption{Offline replay real-time factor (RTF) for specialized memory methods. RTF is processing time to the first output token divided by the duration of the causal video prefix; each entry is the mean over evaluated probes. Values above 1 indicate that processing cannot keep pace with a live 1-FPS video stream under this replay setting. Lower is faster.}
\label{tab:appendix-rtf}
\centering\small
\begin{tabular}{lr}
\toprule
Method & Mean RTF \\
\midrule
HERMES & 0.204 \\
ReKV & 0.089 \\
FluxMem & 0.016 \\
Flash-VStream & 0.010 \\
StreamForest & 0.020 \\
OASIS & 1.223 \\
Video-SALMONN S & 0.010 \\
VST & 0.020 \\
\bottomrule
\end{tabular}
\end{table}

\begin{table}[H]
\caption{Mean end-to-end latency (seconds) for proprietary models, measured from request submission to the complete response. Video memory replays prior sessions; text memory supplies saved summaries and the current causal prefix.}
\label{tab:appendix-api-latency}
\centering\small
\begin{tabular}{lrr}
\toprule
Model & Video memory & Text memory \\
\midrule
Seed-2.0-Lite & 34.11 & 11.25 \\
Gemini 3.6 Flash & 45.73 & 14.58 \\
\bottomrule
\end{tabular}
\end{table}

\subsection{Persistent storage}
\label{app:storage}
Table~\ref{tab:appendix-storage} complements the normalized storage-per-video-hour comparison in \hyperref[sec:evaluation-protocol]{Section~\ref*{sec:evaluation-protocol}} with the mean and maximum prior-session peak state across 104 trajectories. For general video models under \emph{Video as Memory}, the mean and maximum trajectory-peak storage are 2,316.01 and 9,639.01 MiB, respectively; \emph{Text Summary as Memory} occupies only kilobytes. Video-SALMONN S also maintains time-test-training fast weights whose mean trajectory-peak size is 8,404,992 bytes (8.02 MiB). This state is part of the model's parametric adaptation and does not grow with processed video length, so the main efficiency table counts its selected visual memory but does not add those fast weights to per-video storage.

\begin{table}[H]
\caption{Peak persistent-memory size within each trajectory, after completed sessions. Mean and maximum are over 104 trajectories; units are shown in the cells.}
\label{tab:appendix-storage}
\centering\small
\begin{tabular}{lrr}
\toprule
Method & Mean peak & Maximum peak \\
\midrule
HERMES & 330.05 MiB & 330.05 MiB \\
ReKV & 19,202.80 MiB & 58,185.20 MiB \\
FluxMem & 167.31 MiB & 189.96 MiB \\
Flash-VStream & 565.20 MiB & 566.25 MiB \\
StreamForest & 3,374.53 MiB & 3,867.47 MiB \\
OASIS & 2,523.27 MiB & 7,823.59 MiB \\
Video-SALMONN S & 354.34 MiB & 382.91 MiB \\
VST & 4.1 KiB & 6.8 KiB \\
\bottomrule
\end{tabular}
\end{table}

\subsection{Instruction following}
\label{app:invalid}
The automated scorer tolerates format errors that it can repair deterministically. It marks a response incorrect only when no unambiguous task output can be recovered. Examples include an MCQA answer containing multiple options and an Adaptive response that repeats until the maximum output length without a recoverable decision. Tables~\ref{tab:appendix-invalid} and~\ref{tab:appendix-general-invalid} report the remaining failures. Some memory methods show more such failures with long input histories, indicating reduced instruction following under heavy context.

\begin{table}[H]
\caption{Unrecoverable output-format failures for specialized memory methods. Each method is evaluated on 5,371 probes; rate is count divided by 5,371. Recoverable formatting errors are excluded.}
\label{tab:appendix-invalid}
\centering\small
\begin{tabular}{lrr@{\qquad}lrr}
\toprule
Method & Count & Rate & Method & Count & Rate \\
\midrule
HERMES & 43 & 0.80\% & StreamForest & 1,941 & 36.14\% \\
ReKV & 516 & 9.61\% & OASIS & 2 & 0.04\% \\
FluxMem & 199 & 3.71\% & Video-SALMONN S & 792 & 14.75\% \\
Flash-VStream & 492 & 9.16\% & VST & 1,990 & 37.05\% \\
\bottomrule
\end{tabular}
\end{table}

\begin{table}[H]
\caption{Unrecoverable output-format failures for general video models under video and text memory. Counts and percentages use 5,371 probes per setting. SimpleStream uses only the four most recent frames and is reported once.}
\label{tab:appendix-general-invalid}
\centering\small
\begin{tabular}{lrr}
\toprule
Model & Video memory & Text memory \\
\midrule
Seed-2.0-Lite & 14 (0.26\%) & 4 (0.07\%) \\
Gemini 3.6 Flash & 4 (0.07\%) & 3 (0.06\%) \\
Qwen3.8-27B & 27 (0.50\%) & 81 (1.51\%) \\
Qwen3-VL-8B & 15 (0.28\%) & 9 (0.17\%) \\
InternVL3.5-8B & 98 (1.82\%) & 81 (1.51\%) \\
VideoLLaMA3-7B & 875 (16.29\%) & 816 (15.19\%) \\
SimpleStream & \multicolumn{2}{c}{0} \\
\bottomrule
\end{tabular}
\end{table}

\subsection{Effect of trajectory organization}
\label{app:organization}
\begin{figure}[H]
  \centering
  \includegraphics[width=0.8\linewidth]{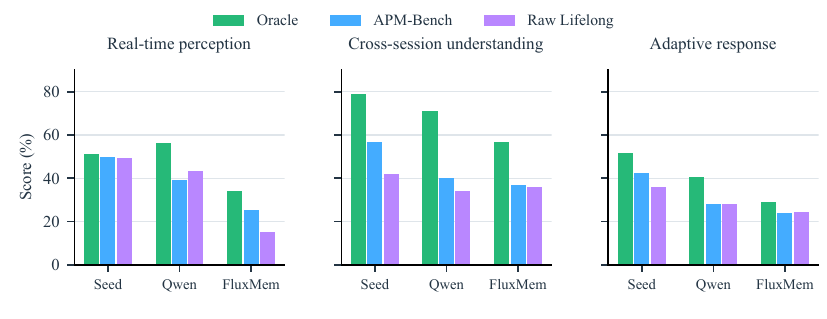}
  \caption{Trajectory organization on 366 candidates. Oracle supplies only necessary evidence; Raw Lifelong adds intervening source video. Adaptive Response uses the \hyperref[sec:evaluation-protocol]{Section~\ref*{sec:evaluation-protocol}} Gated Judge score.}
  \label{fig:organization}
\end{figure}

We compared APM-Bench trajectories with two controls on 12 EgoLife trajectories (366 candidates; 742 runtime instances). \emph{Oracle} supplies the necessary evidence before the cutoff. \emph{Raw Lifelong} adds the recorded video between selected sessions. All conditions use the same tasks and scoring. Figure~\ref{fig:organization} reports results for Seed-2.0-Lite, Qwen3-VL-8B, and FluxMem.

Across these three systems, APM-Bench gives a wider score spread than Raw Lifelong on Cross-session Understanding (standard deviation 8.75 vs.\ 3.33), Adaptive Response (8.03 vs.\ 4.84), and the mean of the three capability scores (8.77 vs.\ 7.05). We compute the population standard deviation across the three model scores as $\sigma=\sqrt{\frac{1}{3}\sum_{m=1}^{3}(s_m-\bar{s})^2}$. On Cross-session Understanding, all three systems also score higher with APM-Bench than with Raw Lifelong: 56.90 vs.\ 41.95, 40.05 vs.\ 34.16, and 37.02 vs.\ 35.97, respectively. Thus, the controlled trajectories make model differences in these two capabilities more visible than the longer raw history in this comparison.

Oracle raises Cross-session Understanding by 19.78--31.21 points over APM-Bench, consistent with relevant-evidence selection being an important difficulty when the history is longer. Adaptive Response gains less under Oracle, and its scores remain 28.98--51.86: access to past evidence alone does not ensure an appropriate response to the current scene. FluxMem's Real-time Perception score falls from 33.95 with Oracle to 25.14 with APM-Bench and 15.28 with Raw Lifelong, as the supplied video grows.

\subsection{Evidence availability and answer accuracy}
\label{app:evidence-availability}
Table~\ref{tab:memory-retention-results} compares Original Accuracy on all 260 questions with Evidence Available Accuracy on the 130 whose evidence remains in the two most recent sessions. Seed-2.0-Lite and Gemini 3.6 Flash score 64.62\% and 85.38\% on this subset, versus 62.69\% and 76.15\% originally. All eight specialized systems score below their original accuracy, including HERMES (7.69\% vs.\ 38.46\%). Accessible evidence alone therefore does not ensure a correct answer. Evidence Unavailable Detection evaluates the other 130 questions; Balanced Accuracy averages the two restricted-history rates.

\begin{table}[H]
  \caption{Evidence Availability-Aware accuracy (\%) on 260 questions. Original Accuracy is the accuracy on the original four-option questions under each system's main evaluation protocol, before restricting history. The restricted-history setting retains the two most recent completed sessions; Balanced Accuracy averages correct answer selection when evidence remains available and insufficient-evidence detection otherwise.}
  \label{tab:memory-retention-results}
  \centering
  \renewcommand{\arraystretch}{1.0}
  \setlength{\tabcolsep}{6pt}
  \resizebox{\textwidth}{!}{%
  \begin{tabular}{lcccc}
    \toprule
    \textbf{Model / Method}
      & \shortstack{\textbf{Original} \\ \textbf{Accuracy}}
      & \shortstack{\textbf{Evidence Available} \\ \textbf{Accuracy}}
      & \shortstack{\textbf{Evidence Unavailable} \\ \textbf{Detection}}
      & \shortstack{\textbf{Balanced} \\ \textbf{Accuracy}} \\
    \midrule
    \rowcolor{gray!15}
    \multicolumn{5}{l}{\textit{\textbf{w/o Memory}}} \\
    SimpleStream & 28.85 & 3.85 & 95.38 & 49.62 \\
    \midrule
    \rowcolor{gray!15}
    \multicolumn{5}{l}{\textit{\textbf{Raw Video as Memory}}} \\
    Seed-2.0-Lite & 62.69 & 64.62 & 66.15 & 65.38 \\
    Gemini 3.6 Flash & 76.15 & 85.38 & 59.23 & 72.31 \\
    Qwen3.8-27B & 58.46 & 59.23 & 46.92 & 53.08 \\
    Qwen3-VL-8B-Instruct & 44.62 & 40.00 & 66.92 & 53.46 \\
    InternVL3.5-8B & 36.15 & 50.77 & 40.77 & 45.77 \\
    VideoLLaMA3-7B & 30.77 & 20.00 & 49.23 & 34.62 \\
    \midrule
    \rowcolor{gray!15}
    \multicolumn{5}{l}{\textit{\textbf{KV Cache}}} \\
    HERMES & 38.46 & 7.69 & 90.77 & 49.23 \\
    ReKV & 34.23 & 10.77 & 63.85 & 37.31 \\
    \midrule
    \rowcolor{gray!15}
    \multicolumn{5}{l}{\textit{\textbf{Visual Tokens / Features}}} \\
    FLUXMem & 37.31 & 15.38 & 71.54 & 43.46 \\
    Flash-VStream & 31.15 & 7.69 & 52.31 & 30.00 \\
    \midrule
    \rowcolor{gray!15}
    \multicolumn{5}{l}{\textit{\textbf{Event Tree}}} \\
    StreamForest & 43.85 & 15.38 & 45.38 & 30.38 \\
    OASIS & 39.23 & 23.08 & 86.15 & 54.62 \\
    \midrule
    \rowcolor{gray!15}
    \multicolumn{5}{l}{\textit{\textbf{Parametric Memory}}} \\
    Video-Salmon-S & 17.31 & 16.15 & 25.38 & 20.77 \\
    \midrule
    \rowcolor{gray!15}
    \multicolumn{5}{l}{\textit{\textbf{Reasoning Thoughts}}} \\
    VST & 24.62 & 10.77 & 55.38 & 33.08 \\
    \bottomrule
  \end{tabular}%
  }
\end{table}

\subsection{Task-level performance}
\label{app:profiles}
\begin{figure}[H]
\centering\includegraphics[width=\linewidth]{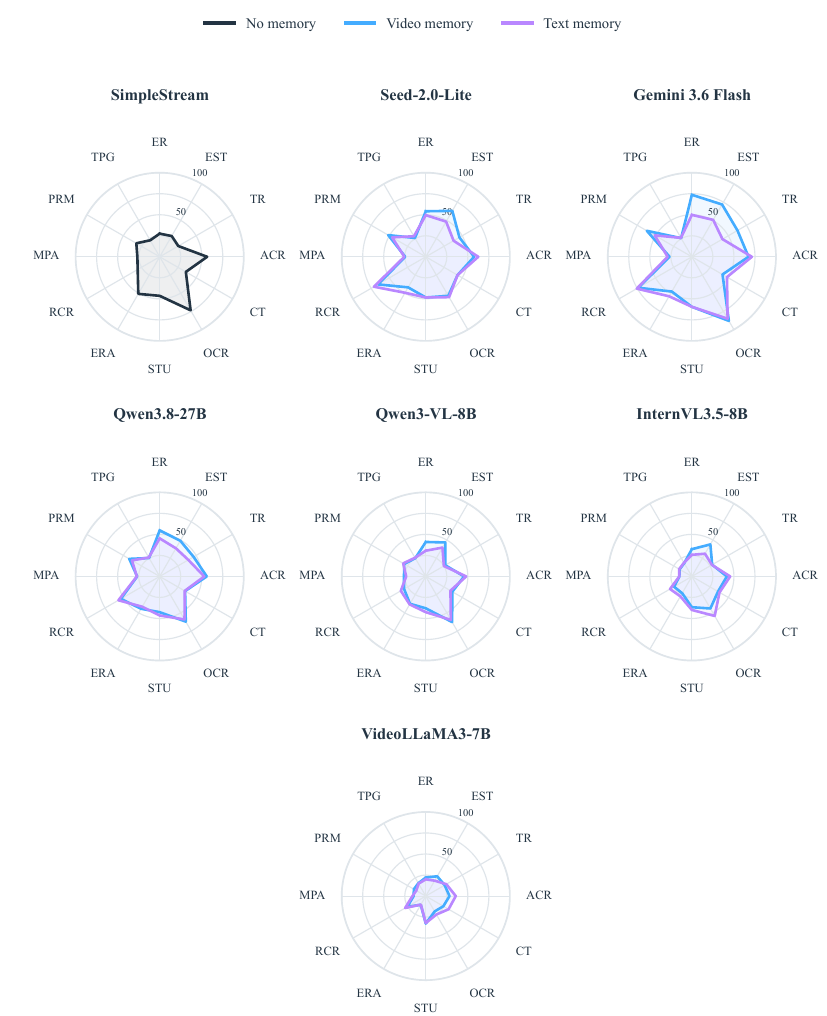}
\caption{Scores on all 12 tasks for six general video models under video- and text-memory protocols. SimpleStream is shown in the same figure for comparison. All axes use a 0--100 scale.}
\label{fig:radar-general}
\end{figure}

\begin{figure}[H]
\centering\includegraphics[width=\linewidth]{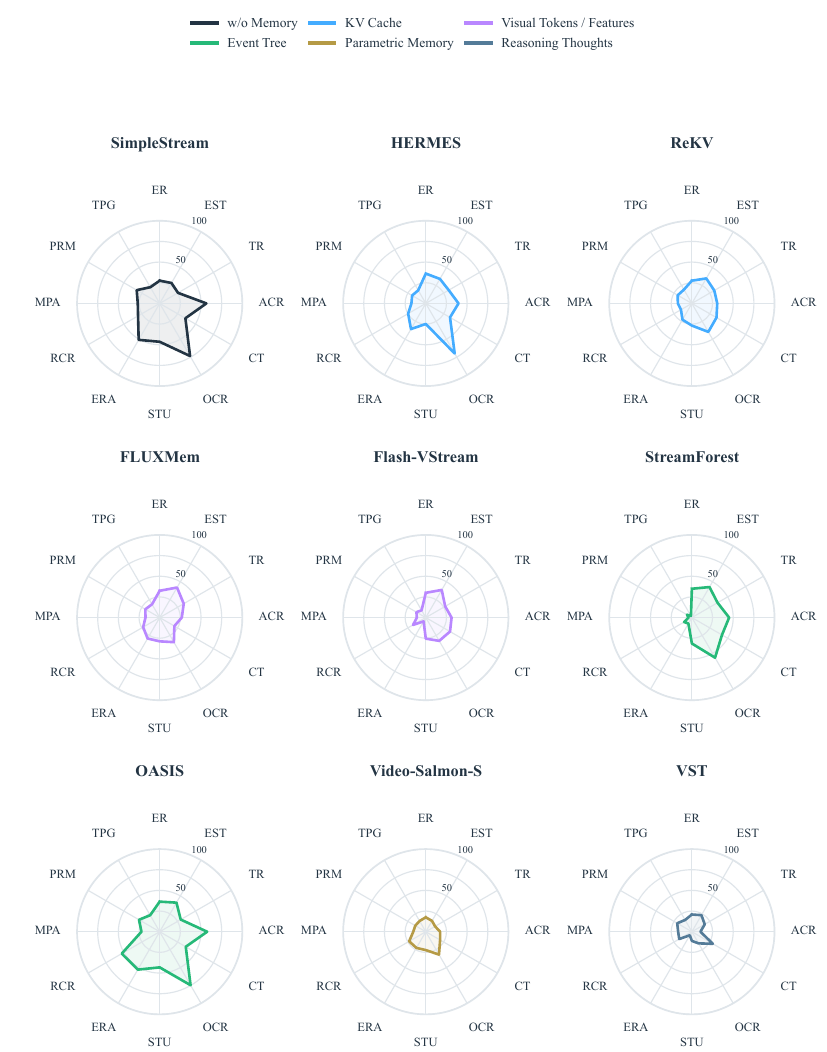}
\caption{Scores on all 12 tasks for eight specialized streaming-memory methods and SimpleStream. All axes use a 0--100 scale.}
\label{fig:radar-specialized}
\end{figure}

Figures~\ref{fig:radar-general} and~\ref{fig:radar-specialized} show the 12 task scores from \hyperref[tab:main-results]{Table~\ref*{tab:main-results}}. The axes group Cross-session Understanding (ER--TR), Real-time Perception (ACR--STU), and Adaptive Response (ERA--TPG). All panels use the same 0--100 scale, and both figures include SimpleStream for comparison.


\clearpage
\section{Dataset Statistics}
\label{app:statistics}

\subsection{Trajectory activities}
\label{app:activities}
\begin{figure}[H]
  \centering
  \includegraphics[width=\linewidth]{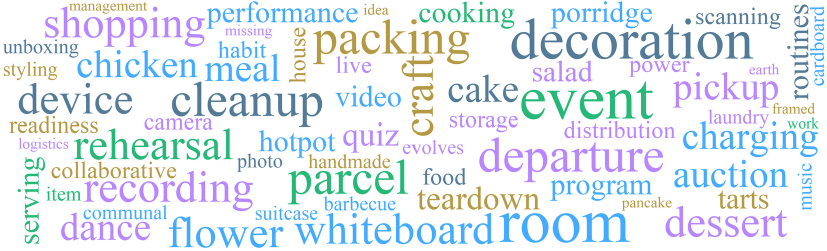}
  \caption{Activity terms in the 104 trajectory titles. Larger words occur in more trajectories.}
  \label{fig:trajectory-wordcloud}
\end{figure}

The 104 trajectories cover recurring everyday activities and longer procedural tasks. Figure~\ref{fig:trajectory-wordcloud} summarizes terms from the curated EgoLife trajectory titles and the HD-EPIC recipe names. Each term is counted at most once per trajectory, so a long title does not dominate the display.

\subsection{Evidence composition}
\label{app:evidence}
Table~\ref{tab:evidence-composition} distinguishes multi-evidence candidates from those whose evidence spans multiple sessions. Multi-evidence means that the answer or response requires at least two distinct evidence items; these items can come from the same session.

\begin{table}[H]
\caption{Evidence composition for inter-session tasks with explicit evidence annotations. Multi-evidence requires at least two evidence items; multi-session evidence spans at least two sessions. Rates use the candidate count in each row. Cross-day means that at least one decisive item precedes the query day.}
\label{tab:evidence-composition}
\centering\setlength{\tabcolsep}{4pt}\small
\begin{tabular}{lccccc}
\toprule
Task & Candidates & Multi-evidence & Multi-session & Same-day & Cross-day \\
\midrule
ER  & 327 & 33 (10.1\%) & 23 (7.0\%) & 53.5\% & 46.5\% \\
EST & 298 & 87 (29.2\%) & 24 (8.1\%) & 55.0\% & 45.0\% \\
TR  & 262 & 262 (100\%) & 173 (66.0\%) & 54.6\% & 45.4\% \\
MPA & 116 & 48 (41.4\%) & 19 (16.4\%) & 59.5\% & 40.5\% \\
TPG & 132 & 73 (55.3\%) & 23 (17.4\%) & 77.3\% & 22.7\% \\
\bottomrule
\end{tabular}
\end{table}

\subsection{Evidence distance measured in sessions}
\label{app:session-distance}
We measure the number of session boundaries between a query and its earliest decisive evidence; for PRM, the earlier registration is the evidence anchor. Table~\ref{tab:session-gap} shows this distance for the 1,249 inter-session candidates. Separately, the decisive evidence occupies one, two, three, four, or five historical sessions for 987, 186, 65, 10, and 1 candidates, respectively.

\begin{table}[H]
\caption{Inter-session candidates by the distance from the query session to the earliest decisive evidence session. Distance one means the immediately preceding session; the last column pools distances of four or more sessions.}
\label{tab:session-gap}
\centering\small
\begin{tabular}{lcccc}
\toprule
Task & 1 session & 2 sessions & 3 sessions & $\geq$4 sessions \\
\midrule
ER & 160 & 80 & 45 & 42 \\
EST & 160 & 78 & 37 & 23 \\
TR & 69 & 74 & 50 & 69 \\
MPA & 63 & 27 & 12 & 14 \\
PRM & 91 & 13 & 3 & 7 \\
TPG & 85 & 26 & 8 & 13 \\
\midrule
\textbf{All} & 628 & 298 & 155 & 168 \\
\bottomrule
\end{tabular}
\end{table}

\subsection{Real-world span and organized trajectory duration}
\label{app:span}
\begin{figure}[H]
  \centering
  \includegraphics[width=\linewidth]{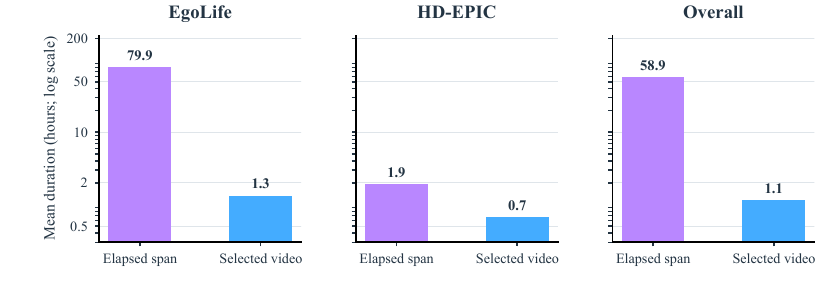}
  \caption{Mean organized trajectory duration and real-world span by source. The span runs from the first session start to the last session end, including gaps; organized duration sums the retained session videos.}
  \label{fig:time-reduction}
\end{figure}

Figure~\ref{fig:time-reduction} compares the organized trajectory duration with its elapsed real-world span, averaged by source and overall. EgoLife retains 1.3 hours of video across a mean span of 79.9 hours; the corresponding values are 0.7 and 1.9 hours for HD-EPIC and 1.1 and 58.9 hours overall. HD-EPIC already contains structured cooking procedures, so organization mainly removes segments unrelated to the recipe and separates the remaining stages into sessions. Its reduction is therefore smaller than EgoLife's.

\section{Data Construction and Annotation Details}
\label{app:data}
\subsection{Dataset construction}
\label{app:dataset-construction}
As shown in Figure~\ref{fig:benchmark-construction}, APM-Bench is constructed in two stages. In \emph{Stage~1}, we organize raw videos into activity-related trajectories. For EgoLife, GPT-5-mini generates hierarchical summaries at hourly and daily scales, and GPT-5.4 mines 165 trajectory proposals from the daily summaries. Human annotators retain 76 valid trajectories comprising 472 sessions. For HD-EPIC, recipe-stage annotations yield 34 trajectory proposals, of which 28 trajectories spanning 77 sessions remain after human verification. For HD-EPIC session videos without visible timestamps, we render real-world timestamps from the source metadata onto the video frames.

In \emph{Stage~2}, Gemini-3.5-Flash generates fine-grained timestamped visual captions for 30-second clips across all sessions. Given the full trajectory captions, DeepSeek-V4-Flash proposes Cross-session Understanding and Adaptive Response candidates; GPT-4o proposes Real-time Perception candidates from selected video frames and captions. The three capabilities yield 6{,}412 initial candidates. We then (1) remove shortcut candidates answered correctly without video by at least two of Qwen3.8-27B, GPT-5, and Gemini-3.1-Pro; (2) discard candidates whose timestamps violate causal constraints; (3) use a tool-augmented VLM agent to audit and refine the remaining candidates; and (4) conduct final human verification and refinement. For 300 questions randomly sampled from the final set, two independent annotators achieved a Cohen's kappa of $0.868$~\citep{cohen1960coefficient}, supporting the reliability of the final annotations.

\subsection{Human verification}
\label{app:verification}
\begin{figure}[t]
  \centering
  \includegraphics[width=\linewidth]{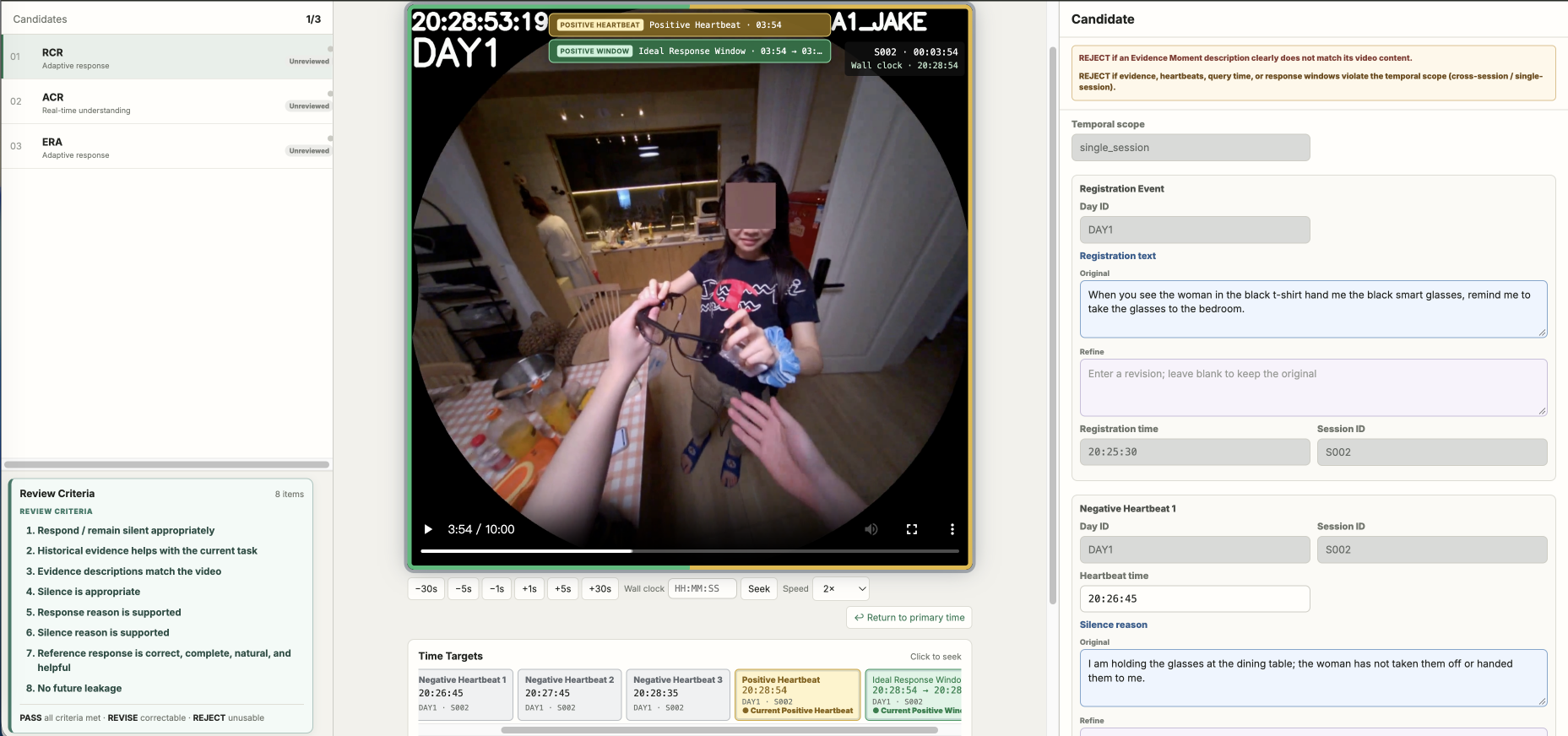}
  \par\smallskip
  \includegraphics[width=\linewidth]{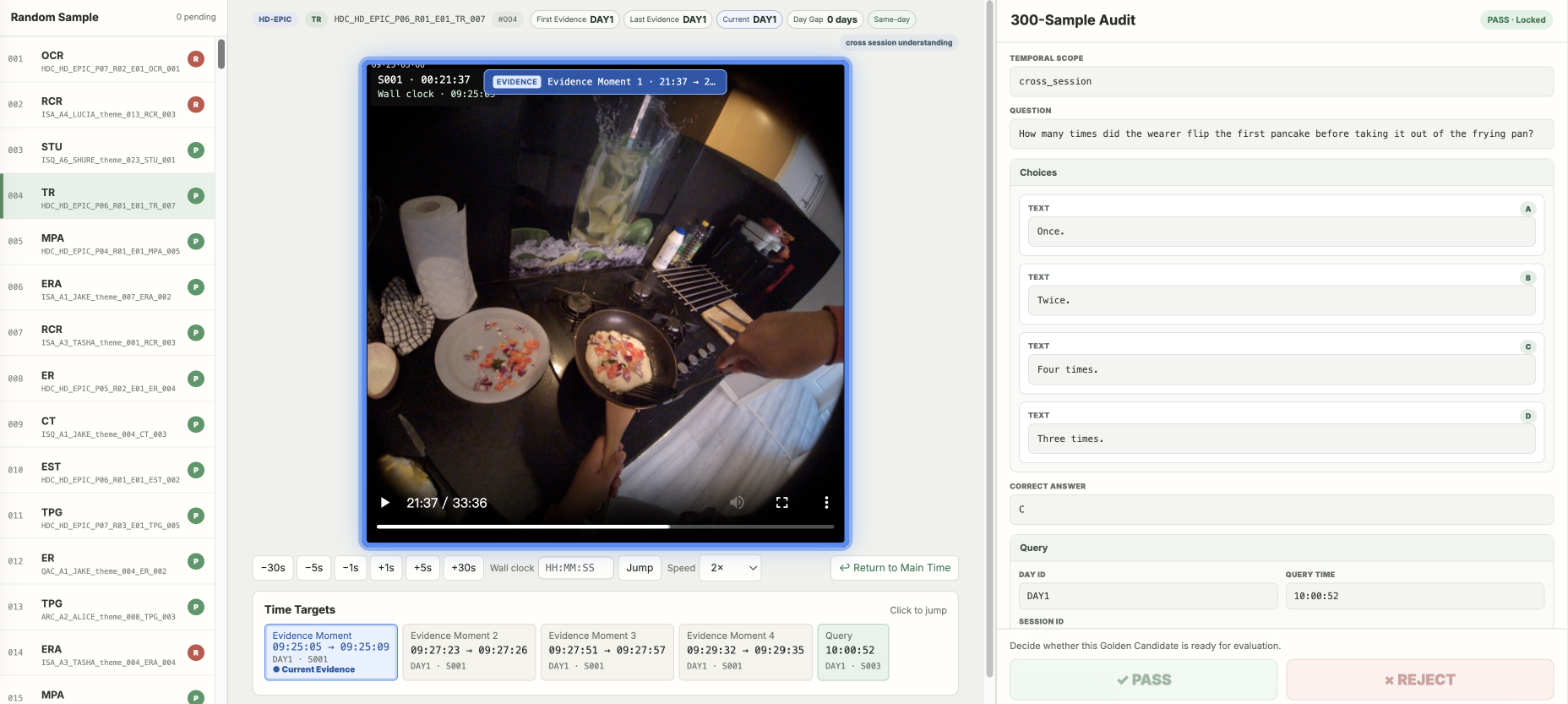}
  \caption{Human review consoles. Top: Human Verify \& Refine shows video, time targets, task criteria, and editable fields for the 3,249 candidates retained after automated screening. Bottom: Agreement Check presents source evidence and candidate fields for the independent 300-question audit.}
  \label{fig:verification-interfaces}
\end{figure}

Reviewers used the Human Verify \& Refine console (Figure~\ref{fig:verification-interfaces}, top) to inspect and revise the 3,249 candidates retained after automated screening. The console places source video and time targets beside the review criteria and editable annotations. For MCQA, reviewers checked question clarity, answer correctness and uniqueness, answerability at query time, evidence support, timestamp accuracy, agreement between evidence descriptions and video, and absence of future leakage. Ambiguous questions were removed.

For Adaptive Response, reviewers checked each probe's intervene/silent label, whether remembered evidence helped with the current task and matched the video, whether silence was justified, and whether response and silence reasons were supported. They checked that reference responses were correct, complete, natural, and useful; verified probe and ideal-response interval timestamps, MPA/TPG historical links, RCR/PRM registrations, and the absence of future leakage; and removed unnecessary interventions or candidates with weak historical links. Reference responses and reasons were refined. The ideal intervals support finer analysis of response timing.

The independent 300-question audit used the Agreement Check console (Figure~\ref{fig:verification-interfaces}, bottom), which displays source evidence alongside each candidate's fields.

Each candidate is tied to a trajectory, task, and causal query point. In Adaptive Response, a positive probe marks an opportunity to respond; a negative probe marks a point at which the assistant should remain silent.

\subsection{MCQA annotation format}
\label{app:candidates}
The seven MCQA tasks (ER, EST, TR, ACR, CT, OCR, and STU) share the same four-option question and evidence structure. The five Adaptive Response annotation formats follow in Sections~\ref{app:era}--\ref{app:tpg}.

\begin{recordbox}{MCQA: ER, EST, TR, ACR, CT, OCR, STU}
{
  "task_type": "ER", "trajectory_id": "...",
  "candidate_id": "...",
  "question": "...",
  "choices": [{"option_id":"A","text":"..."},{"option_id":"B","text":"..."},
              {"option_id":"C","text":"..."},{"option_id":"D","text":"..."}],
  "correct_option_id": "B",
  "query": {"day_id":"DAY4","session_id":"S006",
            "query_time":"18:39:09"},
  "evidence_moments": [
    {"day_id":"DAY1","session_id":"S002",
     "start_time":"20:25:00","end_time":"20:25:30",
     "evidence_content":"..."}
  ]
}
\end{recordbox}

\subsection{ERA: evidence readiness}
\label{app:era}
The four-option question is registered at session start. Subsequent probes mark when its answer becomes available. The released fields \texttt{positive\_heartbeat} and \texttt{negative\_heartbeats} store those probe timestamps.
\begin{recordbox}{ERA: question, evidence, and probe labels}
{
  "task_type":"ERA", "question":"...",
  "choices":[{"option_id":"A","text":"..."},{"option_id":"B","text":"..."},
             {"option_id":"C","text":"..."},{"option_id":"D","text":"..."}],
  "correct_option_id":"D",
  "answer_evidence_moments":[
    {"day_id":"DAY1","session_id":"S002",
     "start_time":"20:32:19","end_time":"20:32:27",
     "evidence_content":"..."}
  ],
  "positive_heartbeat":{"day_id":"DAY1","session_id":"S002",
    "query_time":"20:32:27","response_reason":"..."},
  "negative_heartbeats":[{"day_id":"DAY1","session_id":"S002",
    "query_time":"20:25:45","silence_reason":"..."}],
  "ideal_response_window":{"day_id":"DAY1","session_id":"S002",
    "start_time":"20:32:27","end_time":"20:32:30"}
}
\end{recordbox}

\subsection{RCR: in-session conditional reminder}
\label{app:rcr}
The user registers a condition and response in the current session. Positive and negative probes record when to fulfill the reminder or remain silent; their timestamps appear in the \texttt{heartbeat} fields.
\begin{recordbox}{RCR: registration and probe labels}
{
  "task_type":"RCR",
  "registration_event":{"day_id":"DAY1","session_id":"S001",
    "registration_time":"11:12:30",
    "registration_text":"..."},
  "positive_heartbeat":{"day_id":"DAY1","session_id":"S001",
    "query_time":"11:12:56",
    "proactive_response":"...",
    "reason":"..."},
  "negative_heartbeats":[{"day_id":"DAY1","session_id":"S001",
    "query_time":"11:12:35","silence_reason":"..."}],
  "ideal_response_window":{"day_id":"DAY1","session_id":"S001",
    "start_time":"11:12:55","end_time":"11:12:57"}
}
\end{recordbox}

\subsection{MPA: assistance from prior experience}
\label{app:mpa}
Earlier evidence supports an intervention in the current scene. The record links that evidence to a reference response and separates probes requiring a response from probes requiring silence.
\begin{recordbox}{MPA: historical evidence and probes; probe time = interval end\_time}
{
  "task_type":"MPA",
  "evidence_moments":[{"day_id":"DAY2","session_id":"S001",
    "start_time":"21:43:00","end_time":"21:43:03","evidence_content":"..."},
    {"day_id":"DAY2","session_id":"S001","start_time":"21:43:10",
     "end_time":"21:43:13","evidence_content":"..."}],
  "response_window_rationale":"...",
  "reference_proactive_response":"...",
  "ideal_response_window":{"day_id":"DAY7","session_id":"S003",
    "start_time":"14:24:19","end_time":"14:24:30"},
  "negative_or_silence_windows":[{"day_id":"DAY7","session_id":"S003",
    "start_time":"14:21:00","end_time":"14:21:30","silence_reason":"..."}]
}
\end{recordbox}

\subsection{PRM: cross-session registered reminder}
\label{app:prm}
The earlier registration is paired with a later probe at which the reminder becomes due and probes at which it should remain silent.
\begin{recordbox}{PRM: registration and probes; probe time = interval end\_time}
{
  "task_type":"PRM",
  "registration_event":{"day_id":"DAY5","session_id":"S005",
    "registration_time":"20:55:30",
    "registration_text":"..."},
  "trigger_match_reason":"...",
  "reference_proactive_response":"...",
  "ideal_response_window":{"day_id":"DAY6","session_id":"S008",
    "start_time":"16:24:26","end_time":"16:24:27"},
  "negative_or_silence_windows":[{"day_id":"DAY6","session_id":"S008",
    "start_time":"16:23:30","end_time":"16:24:00","silence_reason":"..."}]
}
\end{recordbox}

\subsection{TPG: continuing-task guidance}
\label{app:tpg}
Historical evidence and the current scene identify when guidance for the continuing task is useful. The record stores the reference response and the reasons for intervening or remaining silent.
\begin{recordbox}{TPG: continuing-task evidence and probes; probe time = interval end\_time}
{
  "task_type":"TPG",
  "evidence_moments":[{"day_id":"DAY2","session_id":"S003",
    "start_time":"13:11:16","end_time":"13:11:23","evidence_content":"..."},
    {"day_id":"DAY2","session_id":"S005","start_time":"16:37:30",
     "end_time":"16:38:00","evidence_content":"..."}],
  "memory_relevance":"...",
  "ideal_response_windows":[{"day_id":"DAY5","session_id":"S007",
    "start_time":"23:01:28","end_time":"23:01:32",
    "response_window_rationale":"...",
    "reference_proactive_response":"..."}],
  "negative_or_silence_windows":[{"day_id":"DAY5","session_id":"S007",
    "start_time":"23:00:02","end_time":"23:00:30","silence_reason":"..."}]
}
\end{recordbox}

\section{Task Prompts}
\label{app:prompts}
Angle-bracketed fields in the templates below are filled from the candidate or probe. Box titles and the two ERA dividers mark separate runtime calls; they are not part of the prompt text.

\subsection{System prompt}
\label{app:system-prompt}
The shared instruction precedes every task prompt and restricts the assistant to information available in the causal stream.

\begin{promptbox}{Shared system instruction}
You are a first-person streaming video assistant.

Use only the visual stream, persistent memory, and user interactions made available to you. Do not assume access to future video or to information that has not been provided.

Return exactly one valid JSON object in the requested format. Do not add markdown or text outside the JSON object.
\end{promptbox}

\subsection{MCQA prompts}
\label{app:mcqa-prompt}
ER, EST, TR, ACR, CT, and OCR use the four-option template below. STU uses the same answer format with an additional instruction to interpret spatial relations from the camera wearer's viewpoint.

\begin{promptbox}{MCQA: ER, EST, TR, ACR, CT, OCR}
Based only on the information available up to the current moment, answer the following multiple-choice question.


Question:
<QUESTION>

Choices:
A. <OPTION A>
B. <OPTION B>
C. <OPTION C>
D. <OPTION D>

Return exactly:
{"answer": "A|B|C|D"}
\end{promptbox}

\begin{promptbox}{STU: egocentric viewpoint guidance}
Based only on the information available up to the current moment, answer the following multiple-choice question.

Task guidance:
Interpret left, right, front, behind, and other viewpoint-dependent directions from the camera wearer's egocentric viewpoint unless the question explicitly defines another reference frame. For object-to-object relations, use the reference object stated in the question.

Question:
<QUESTION>

Choices:
A. <OPTION A>
B. <OPTION B>
C. <OPTION C>
D. <OPTION D>

Return exactly:
{"answer": "A|B|C|D"}
\end{promptbox}

\subsection{Adaptive Response probes}
\label{app:adaptive-prompt}
The five Adaptive tasks use separate probes. ERA first registers a question at session start; the two labeled parts of its box are sent at different times. RCR likewise receives the user's registration before its probe. For MPA, PRM, and TPG, an unlabeled interval starts with the marker shown below, and the probe occurs at its end time. The original template wording calls ERA and RCR probes ``checkpoints'' and the other probes ``intervals.''

\begin{promptbox}{ERA: question registration and response probe}
Session-start question:
At the beginning of this session, the user asked a delayed-answer question. Do not guess or answer it immediately; retain it and wait for a response checkpoint.

Question:
<QUESTION>

Choices:
A. <OPTION A>
B. <OPTION B>
C. <OPTION C>
D. <OPTION D>

Response probe:
This is a response checkpoint. Based only on the causal information available at this checkpoint, decide whether the previously asked question now has one reliable and uniquely determined answer. Do not guess or use a result that is only revealed later.

If the answer is not yet uniquely determined, return:
{"decision": "SILENT", "reason": "..."}

If the answer is now uniquely determined, return:
{"decision": "INTERVENE", "answer": "A|B|C|D", "reason": "..."}
\end{promptbox}

\begin{promptbox}{RCR: response probe}
This is a response checkpoint. Decide whether a reminder obligation previously registered by the user is visually triggered at this checkpoint and should be fulfilled now.

Use the registered condition and requested response together with the causal visual evidence. The existence of a registration alone is not a trigger. Intervene only when the registered condition is currently observable and satisfied.

Do not assume that a response from another evaluation call has already been delivered. Remain silent when the registered condition is not currently satisfied or the visual evidence is insufficient.

If no response should be provided, return:
{"decision": "SILENT", "reason": "..."}

If a registered response should be provided, return:
{"decision": "INTERVENE", "proactive_response": "...", "reason": "..."}
\end{promptbox}

\begin{promptbox}{Interval-start marker for MPA, PRM, and TPG}
The unlabeled evaluation interval begins now at <START_TIME>. Assess assistance only from this marker through the current cutoff.
\end{promptbox}

\begin{promptbox}{MPA: probe at visual-interval end}
Evaluate the following unlabeled half-open interval in the causal visual stream:
[<START_TIME>, <END_TIME>)

Decide whether useful and timely proactive assistance should be provided during that interval. Use earlier experiences, preferences, or repeated behavior only when they materially improve concrete assistance in the current situation.

Intervene only when the currently observable situation makes a response useful and actionable. Remain silent when the evidence is insufficient, the relevant condition has not been met, or the opportunity is not currently actionable.

If no response should be provided, return:
{"decision": "SILENT", "reason": "..."}

If a response should be provided, return:
{"decision": "INTERVENE", "proactive_response": "...", "reason": "..."}
\end{promptbox}

\begin{promptbox}{PRM: probe at visual-interval end}
Evaluate the following unlabeled half-open interval in the causal visual stream:
[<START_TIME>, <END_TIME>)

Decide whether useful and timely proactive assistance should be provided during that interval. Consider whether a previously registered reminder obligation is visually triggered during this interval and should be fulfilled. Use the registered condition and requested response together with the causal visual evidence. The existence of a registration alone is not a trigger. Intervene only when the registered condition is currently observable and satisfied.

Intervene only when the currently observable situation makes a response useful and actionable. Remain silent when the evidence is insufficient, the relevant condition has not been met, or the opportunity is not currently actionable.

If no response should be provided, return:
{"decision": "SILENT", "reason": "..."}

If a response should be provided, return:
{"decision": "INTERVENE", "proactive_response": "...", "reason": "..."}
\end{promptbox}

\begin{promptbox}{TPG: probe at visual-interval end}
Evaluate the following unlabeled half-open interval in the causal visual stream:
[<START_TIME>, <END_TIME>)

Decide whether useful and timely proactive assistance should be provided during that interval. Consider whether progress in the same ongoing longer-term task makes stage-appropriate guidance useful now.

Intervene only when the currently observable situation makes a response useful and actionable. Remain silent when the evidence is insufficient, the relevant condition has not been met, or the opportunity is not currently actionable.

If no response should be provided, return:
{"decision": "SILENT", "reason": "..."}

If a response should be provided, return:
{"decision": "INTERVENE", "proactive_response": "...", "reason": "..."}
\end{promptbox}

\subsection{Evidence Availability-Aware prompt}
\label{app:evidence-availability-prompt}
The Evidence Availability-Aware evaluation gives models access only to the two most recent completed sessions. Its fifth option asks the model to identify questions whose required evidence is outside that accessible history.

\begin{promptbox}{Evidence Availability-Aware MCQA: evidence-sufficiency variant}
Based only on the causal information available up to the current moment, first determine whether the evidence is sufficient to support one reliable answer and whether this is the appropriate time to answer.


If the evidence is sufficient, select the best content answer. If the evidence is insufficient or answering would require information revealed only later, select the option that explicitly states it is not yet appropriate to answer.

Question:
<QUESTION>

Choices:
A. <OPTION A>
B. <OPTION B>
C. <OPTION C>
D. <OPTION D>
E. <OPTION E>

Return exactly:
{"answer": "A|B|C|D|E"}
\end{promptbox}

\subsection{Text-summary memory prompt}
\label{app:text-summary-prompt}
At each session end, the memory writer produces a summary without access to future questions. Summaries from completed sessions are supplied with the current causal video prefix at later queries or Adaptive probes. The writer uses the following instruction and session-end request.

\begin{promptbox}{Text-summary protocol: memory-writer instruction}
You maintain query-agnostic persistent memory for a first-person streaming video assistant.

Observe this complete session and the user interactions that occur during it. After the session ends, write a detailed, self-contained memory of information that may remain useful in later sessions. The original session will not be available then.

Decide what to retain and how to organize it without anticipating any future evaluation question. Preserve events, entities, states, locations, task progress, corrections, preferences, and registered obligations when they are visually or explicitly supported. Do not invent details; preserve uncertainty when needed.

Return exactly one valid JSON object in this format:
{"memory": "A detailed, self-contained memory of the session."}
\end{promptbox}

\begin{promptbox}{Text-summary protocol: session-end request}
Session metadata:
- day: <DAY_ID>
- session: <SESSION_ID>
- wall-clock interval: [<START_TIME>, <END_TIME>)
The session has ended. Write its persistent memory now.
\end{promptbox}

\end{document}